\documentclass{frontiersSCNSmod}

\usepackage{url,hyperref,lineno,microtype,subcaption}
\usepackage[onehalfspacing]{setspace}
\usepackage{csquotes}
\usepackage[colorinlistoftodos]{todonotes}
\usepackage{color}

\def\keyFont{\fontsize{8}{11}\helveticabold }
\def\firstAuthorLast{Pilarski {et~al.}}
\def\Authors{Patrick M. Pilarski\,$^{1,2*}$, Richard S. Sutton\,$^{2}$,  Kory W. Mathewson\,$^{1,2}$,\\Craig Sherstan\,$^{1,2}$, Adam S. R. Parker\,$^{1,2}$, and Ann L. Edwards\,$^{1,2}$}
\def\Address{$^{1}$Division of Physical Medicine and Rehabilitation, Department of Medicine, University of Alberta, Edmonton, AB, Canada\\
$^{2}$Reinforcement Learning and Artificial intelligence Laboratory, Department of Computing Science, University of Alberta, Edmonton, AB, Canada}
\def\corrAuthor{Patrick M. Pilarski, Division of Physical Medicine and Rehabilitation, Department of Medicine, 5-005 Katz Group Centre for Pharmacy and Health Research, University of Alberta, Edmonton, AB, Canada, T6G 2E1.}

\def\corrEmail{pilarski@ualberta.ca}

\newcommand{\etal}{\mbox{\emph{et al.}}}

\begin{document}
\onecolumn
\firstpage{1}

\title[Communicative Capital for Prosthetic Agents]{Communicative Capital for Prosthetic Agents} 

\author[\firstAuthorLast ]{\Authors}
\address{}
\correspondance{}

\extraAuth{}

\maketitle

\begin{abstract}

\section{}
This work presents an overarching perspective on the role that machine intelligence can play in enhancing human abilities, especially those that have been diminished due to injury or illness. As a primary contribution, we develop the hypothesis that assistive devices, and specifically artificial arms and hands, can and should be viewed as agents in order for us to most effectively improve their collaboration with their human users. We believe that increased agency will enable more powerful interactions between human users and next generation prosthetic devices, especially when the sensorimotor space of the prosthetic technology greatly exceeds the conventional control and communication channels available to a prosthetic user. To more concretely examine an agency-based view on prosthetic devices, we propose a new schema for interpreting the capacity of a human-machine collaboration as a function of both the human's and machine's degrees of agency. We then introduce the idea of {\em communicative capital} as a way of thinking about the communication resources developed by a human and a machine during their ongoing interaction. Using this schema of agency and capacity, we examine the benefits and disadvantages of increasing the agency of a prosthetic limb. To do so, we present an analysis of examples from the literature where building communicative capital has enabled a progression of fruitful, task-directed interactions between prostheses and their human users. We then describe further work that is needed to concretely evaluate the hypothesis that prostheses are best thought of as agents. The agent-based viewpoint developed in this article significantly extends current thinking on how best to support the natural, functional use of increasingly complex prosthetic enhancements, and opens the door for more powerful interactions between humans and their assistive technologies.

\tiny
 \keyFont{ \section{Keywords:} machine intelligence, prosthetics, human-machine interaction, agency, prediction learning, communication, rehabilitation technology, robotic systems}
\end{abstract}

{\bf WORD COUNT}: 11,995 words (281 in abstract)

\section{Introduction}

Technology is routinely used to enhance human abilities. Notable examples include the computers that sit on most desks, smart devices in our pockets, and  medical technology affixed to or implanted within our bodies. All of these enhance human potential in one or more ways. We use the term {\em enhancement} to refer to any process or procedure which adds to the functional capabilities of an individual; enhancement may include amplifying natural abilities and replacing natural abilities that were lost (e.g., through injury or illness). Examples of enhancement are prominent in both academic study and in mainstream culture. This prominence is illustrated by numerous real-world examples (e.g., Geary 2002, Doidge 2007, Dewdney 1998, Brooks 2002, Belfiore 2009, and Moss 2011) and in creative fiction by authors like William Gibson, Neil Stephenson, and Max Barry.

Some technologies are used for improving our ability to control or influence the world ({\em motor enhancement}),  some for improving our perceptual abilities ({\em sensory enhancement}), and some for altering our thought and memory ({\em cognitive enhancement}). Enhancement in these three areas is something that ordinary people do (Clark 2008), and also something that is used to meet specific needs for specific individuals---e.g., assistive devices for people that have lost limbs or other capacities. In other words, these three forms of enhancement are normal and continuous with ordinary things humans do every day and have done since humans developed the capacity for tool use. These forms of enhancement are also significant (and in some cases controversial)  areas of active research and development in academia and industry. As notable examples, Doidge (2007) describes ways that technology can leverage brain plasticity to improve cognitive function in subjects with healthy and injured brains; Geary (2002) describes ways that technology is currently used to enhance sight, touch, hearing, taste, smell, and even mental processes; and Mill\'an \etal\ (2010), Castellini \etal\ (2014), and Carmena (2012) all present views into the use of technology to restore motor and sensory abilities to people who have lost body parts or body functions. The goal of the present article is to develop an over-arching perspective on how human abilities can be enhanced using one specific class of technologies known as {\em machine intelligence}---a machine's independent pursuit of knowledge and use of this knowledge in decision making.

\section{Robotic upper-limb prostheses}

Enhancing technologies take many forms, ranging from ordinary to extraordinary, technological to biological, and have incredible potential for economic and social impact. Because of this diversity it can be challenging to take the lessons learned from one kind of interaction and apply them to a  different enhancement technology. In the present work we seek to identify important commonalities in the different forms of enhancement. Having found these similarities, we hope to leverage them such that each individual application or technology does not need to be treated separately with different methods and different approaches. 

To focus our thinking on areas that can be well supported by machine intelligence, we  consider highly technological means of enhancement wherein information processing and computation are important aspects of a device's capacity for assistance and augmentation. In particular, we examine the setting of assistive rehabilitation technology. 
The setting of assistive rehabilitation technology is appealing in that it involves a direct, immediate collaboration between a human and their technology (here termed a {\em device}) to achieve a goal. Examples of advanced assistive devices intended to support human activities include semi-autonomous wheelchairs (Mill\'an \etal\ 2010, Viswanathan \etal\ 2014), exoskeletons for both paraplegics and their care givers (Herr 2009), smart living environments (Rashidi and Mihailidis 2013), and socially assistive robotic coaches that provide therapeutic support (Feil-Seifer and Matari\'c 2011). There are also biological examples of assistance that can be thought of in a similar way to non-biological assistive technologies. For example, seeing eye dogs have been used for intelligent, decision making assistive roles as far back as the 13th century (Fishman 2003).

\begin{figure}[t]
\centering
\includegraphics[height=3in]{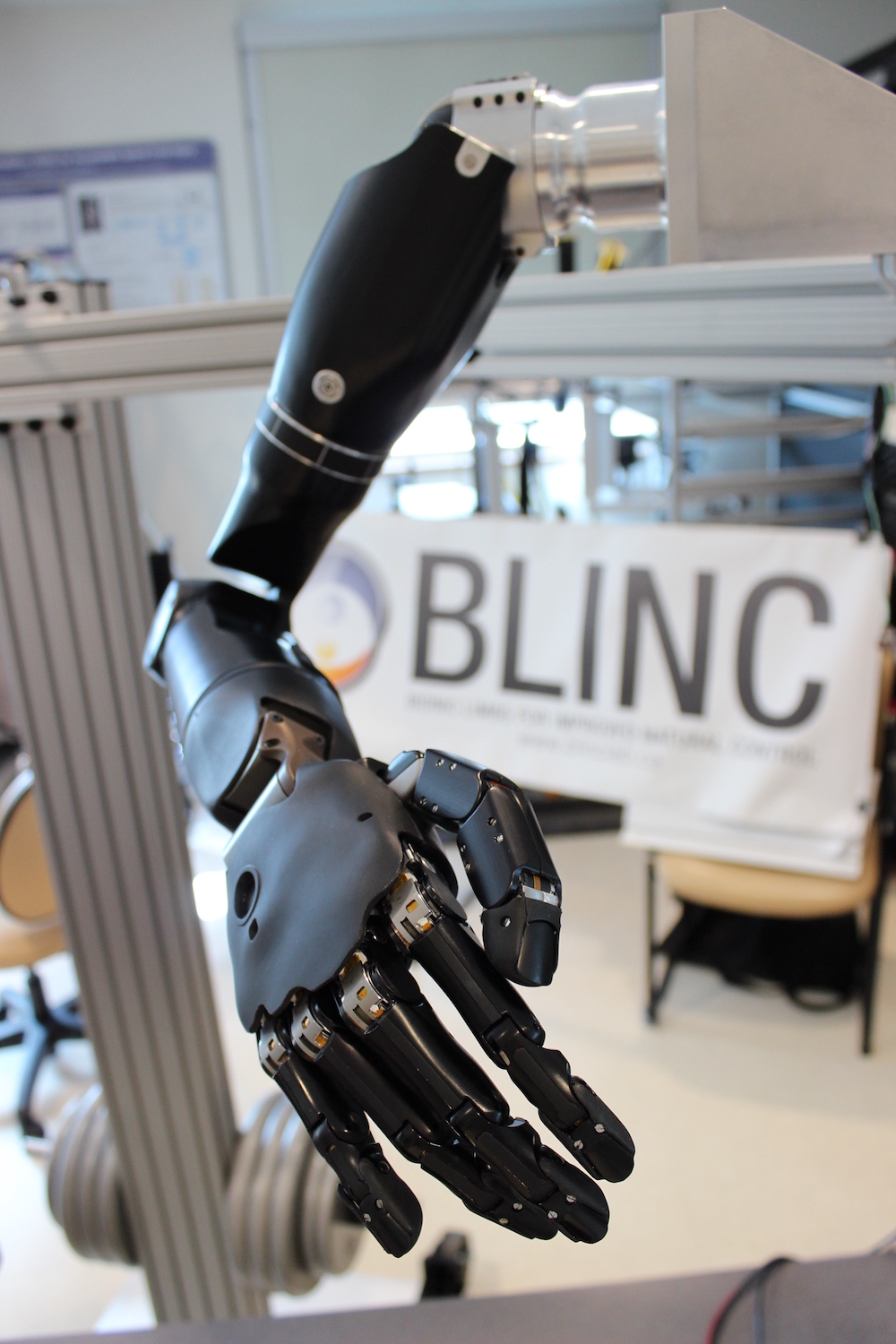}
\includegraphics[height=3in]{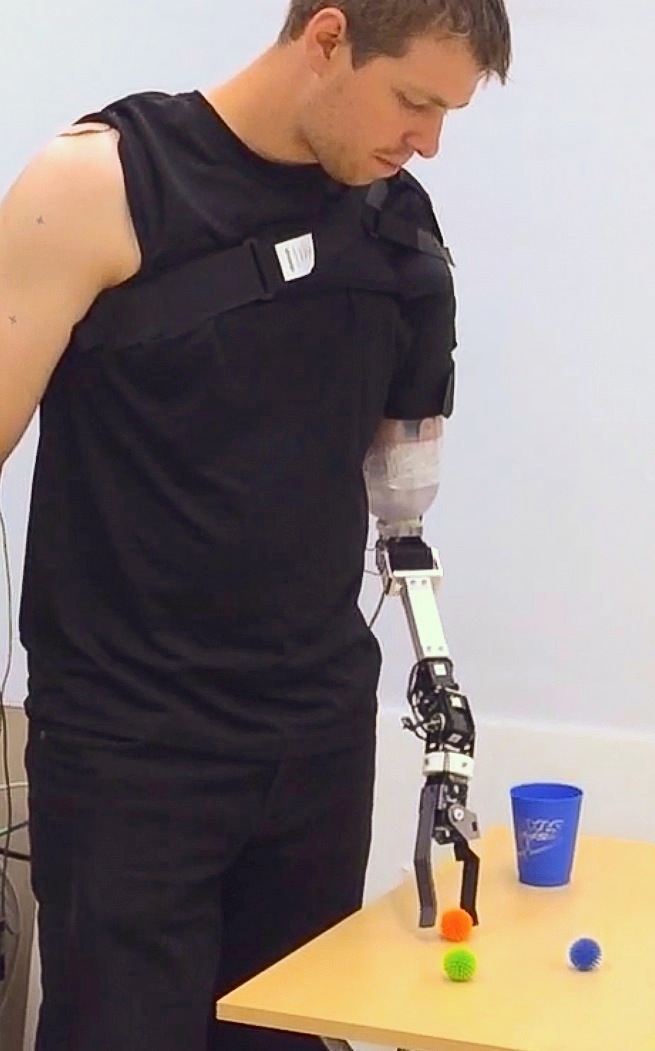}
\includegraphics[height=3in]{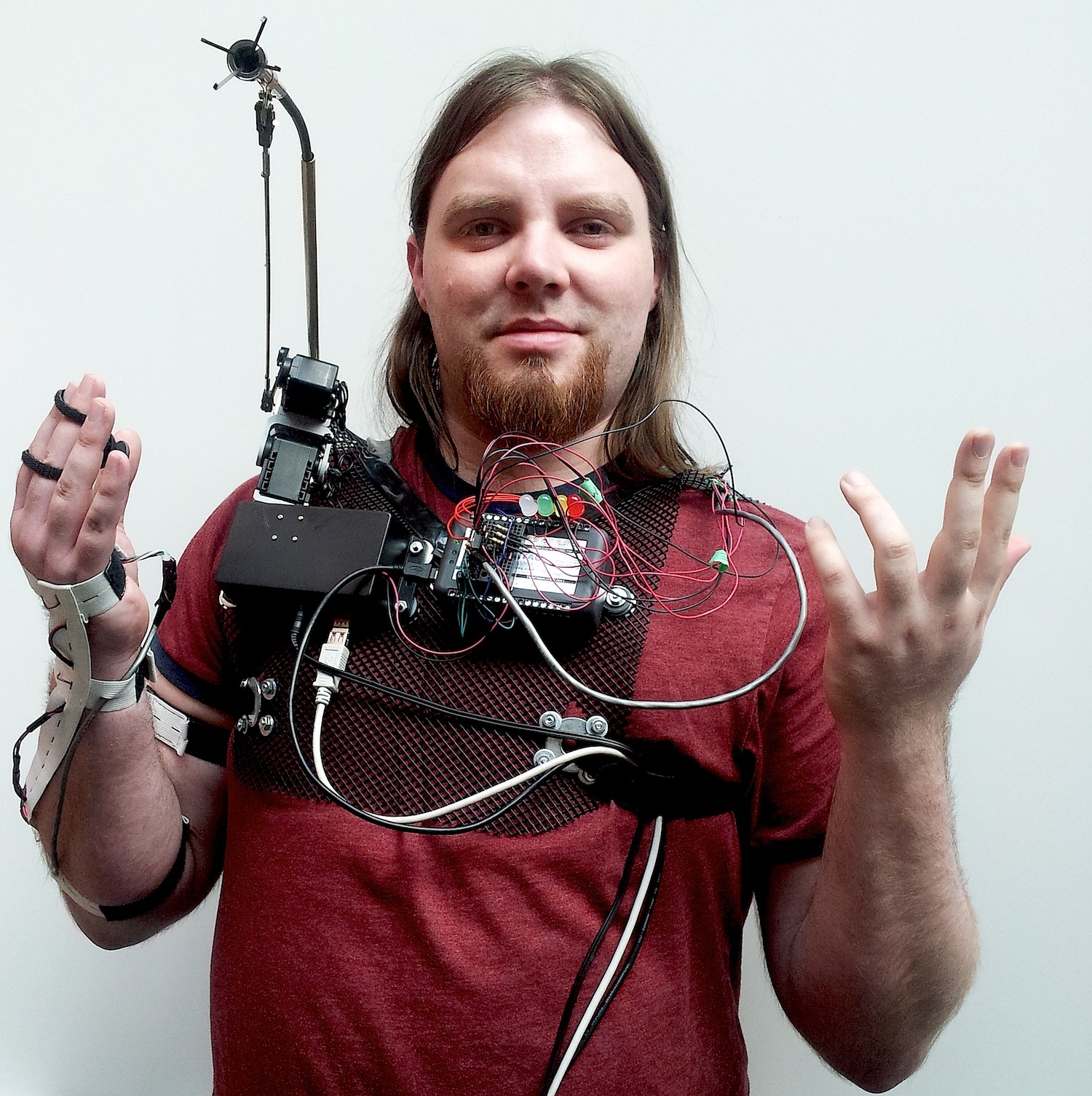}\\
\bf (a) \hfil (b) \hfil (c)
\caption{Examples of prostheses and human-prosthesis interaction. (a) an advanced prosthetic device with over 20 degrees-of-freedom, on-board computer, and numerous reportable sensory precepts. (b) a subject with an amputation using the University of Alberta Bento Arm (Dawson \etal\ 2014) with conventional myoelectric control to complete a manipulation task. (c) control of a supernumerary limb by a non-amputee subject (Parker \etal\ 2014).}
\label{fig:examples}
\end{figure}

A representative example of assistive rehabilitation technology, and the one we will focus on for the remainder of this manuscript, is that of {\em \nobreak robotic upper-limb prostheses}: electromechanical devices attached to the body of a user during daily life to augment, replace, or in other ways restore hand and arm function lost due to the absence of a biological limb (e.g., from amputation, injury, or illness) (Fig. \ref{fig:examples}). Upper-limb prosthetic devices have evolved over the last several hundred years from crude iron hands to exquisitely designed bionic body parts (Zuo and Olson 2014, Castellini \etal\ 2014). However, despite great improvements in quality of life for those with lost limbs, the state-of-the-art has yet to create a satisfactory substitute for the nearly 1 in 200 Americans living with amputations (Zuo and Olson 2014; Castellini \etal\ 2014, Ziegler-Graham \etal\ 2008, Peerdeman \etal\ 2011). Significant effort is being expended by both academic researchers and industry members to bridge the gap between users and their increasingly complex bionic body parts (Williams 2011).

From a technical standpoint, the prosthetic setting is  both challenging and appealing due to the intimate way interactions with the environment are shared between a human and their device. The user of a prosthesis must treat the limb as part of their own body despite limitations in their control of the limb's actuators and a shortage of received feedback from the limb (Casellini \etal\ 2014, Schofield \etal\ 2014). Technically sound coupling of a user to their device is further complicated by the dynamic, non-stationary nature of human environments (Saridis and Stephanou 1977), and by a radical increase in the amount of data present in the human-machine interface due to the development of supporting technologies. Muscular, neural, and osseointegration are now allowing more direct, high-bandwidth  connections between the human and the device (Zuo and Olson 2014, Castellini \etal\ 2014, Hochberg \etal\ 2006, Oritz-Catalan \etal\ 2014). Cameras have been recently implemented in a research setting to inform prosthesis control (Markovic \etal\ 2014), microphones and speakers have been deployed to facilitate natural-language interactions with devices in other robotic domains (Kollar \etal\ 2010), and both surgical practices and prosthetic feedback approaches have evolved to provide a more advanced, bidirectional flow of data between prostheses and their users  (Schofield \etal\ 2014, Hebert \etal\ 2014). Future prosthetic devices will receive an unprecedented density of data about the user, their needs, and their environment.
In this work, we suggest that prosthetic devices will need to take an active, ongoing role in structuring and acting upon the information contained in this data to best support their users' needs and goals.

\section{Hypothesis: Prostheses as Agents}

We propose that by increasing the autonomy of a prosthetic device one can dramatically increase the capabilities of the full human-prosthesis partnership. The principal contribution of this work is to explore the hypothesis that a prosthetic device should be an {\em agent}---i.e., that it is useful to consider a  prosthesis to be an autonomous goal-seeking system. In more general terms, we propose that the parts of a larger information processing system (e.g., both sides of a tightly coupled human-machine interface) are well thought of as each being full information-processing systems, {\em agents}, with goals. We further suggest that, for maximum benefit, all parts of an interface should take into account and model the agency of the other parts. As a foundation for this approach we build on the idea of {\em communicative capital}---a set of communication resources akin to a language that may built up through ongoing interactions between a human and their machine counterpart. \footnote{The ideas of communicative capital and the goal-seeking agency of prosthetic devices were first introduced explicitly in a workshop paper by Pilarski \etal\ (2015). These ideas are more completely presented and significantly extended in the current work.}

 In the remainder of this manuscript we will present an intuition behind an agent-based viewpoint and the development of communicative capital that supports it. To do so, we put our viewpoint into context using recent examples of how ongoing machine learning has been used to improve the operation of powered prosthetic limbs and other assistive robots. We close with an outline of the steps required for rigorously testing of our hypothesis and putting the subsequent conclusions into practice. The result is a new theory about how scientists, engineers, and medical professionals can work together to best pursue human enhancement in the face of injury or illness.

\section{Agency: Having and Seeking Goals}

We now define a schema for thinking about the levels of agency and the resulting capabilities that each side of the human-machine interface may obtain. In what follows, we will refer to the human as the {\em director}, and their prosthesis, or other assistive device, as the {\em assistant}. The director and the assistant may be well thought of as co-actors in a joint action task, as described by Sebanz \etal\ (2006) and Pezzulo \etal\ (2013), or the leader and follower in a two-agent partnership (Candidi \etal\ 2015). For the purposes of our present discussion, we define agency as the degree to which an autonomous system has the ability to have, seek, and achieve goals. This definition is similar to the Belmont Report (1979), wherein a system assumes agency if it is ``capable of deliberation about personal goals and of acting under the direction of such deliberation". Common hallmarks of agency include the ability to take actions, have sensation, persist over time, and improve with respect to a goal; these hallmarks give rise to an agent's ability to predict, control, and model its environment (including other agents). As one example, Tosic \etal\ (2004) explore a hierarchical taxonomy of autonomous agents by defining weak and strong autonomy and exploring combinations of characteristics of autonomy. They describe weakly autonomous agents as having attributes such as control over one's own state, being reactive to stimulus, and persisting over time, while strong autonomy includes attributes such as goal-driven behaviour, being proactive, and being able to adapt. While these attributes may not fully encapsulate agency, they provide a framework of building blocks to discuss higher level, more complex autonomous agents.

Taking prior perspectives on agency into consideration, along with the nuances of the prosthetic setting of interest, we choose to focus here on five attributes or types of agency that may be present in different combinations in a director and/or assistant. These attributes, while certainly not comprehensive, provide good coverage of the space of examples relevant to our analysis of human-prosthesis interaction. Thus, for the purpose of the discussions that follow, a director or assistant may:

{\bf Be a mechanism:} The system acts in a fixed or predetermined way in response to the state or stimulus. For example, the standard case of a body-powered prosthesis, or a conventional myoelectric controller that processes EMG signals via a fixed linear proportional mapping to create control commands for the prosthetic actuators (Parker \etal\ 2006). In the simplest prosthetic control cases, a mechanism may be considered to be an open-loop controller with an immutable mapping from inputs to outputs.

{\bf Adapt over time:} The system is a changing mechanism. In addition to acting mechanistically, the system has the capacity to adapt or change in a persistent way in response to the situation and signals perceived from the other agent. One clear way of thinking of this case is a system that gradually acquires knowledge about its situation in the form of changing parameters, thresholds, or forecasts about how to act and what will happen in the future (i.e., prediction and control learning). Adaptation can occur over scheduled periods of time, as in the supervised learning of a pattern recognition classifier, or during ongoing experience (Castellini \etal\ 2014, Pilarski \etal\ 2013a).

{\bf Have a goal:} The system has defined goals, with the intent to maximize or optimize some measure of its own situation. One way that goals may be defined is in the form of scalar signals of reward (success), as in the computational and biological reinforcement learning literature (Sutton and Barto 1998). This level of agency is the common case for the director---the human user of a commercially available myoelectric or mechanical prosthesis.

{\bf Model or think of the other agent as adapting:} The agent views the other agent as changing and building up expectations (for example, predictions) during ongoing interaction, and in response to the signals it generates. This alters the way the first agent presents signals to the second agent. An example of this level of interaction is an amputee (the director) training a pattern recognition controller (the assistant), knowing that the assistant is adapting to the signals the director generates.  

{\bf Model or think of the other agent as having a goal:} The agent views the other agent as not only changing in response to received signals, but also as having its own objectives. This may be viewed as the agent having at least a preliminary ``theory of mind,'' further altering the way one agent presents signals to the other agent. Viewing another agent as an adaptive, goal-seeking system enables more advanced forms of direction, collaboration, and instruction.

As is evident even in the selected views on agency presented above, agency is not easily identified as present or absent in a non-human system. Taken further, one can imagine additional attributes of agency that begin to approach a full theory of mind.

\begin{figure}[t]
\centering
\includegraphics[height=3.5in]{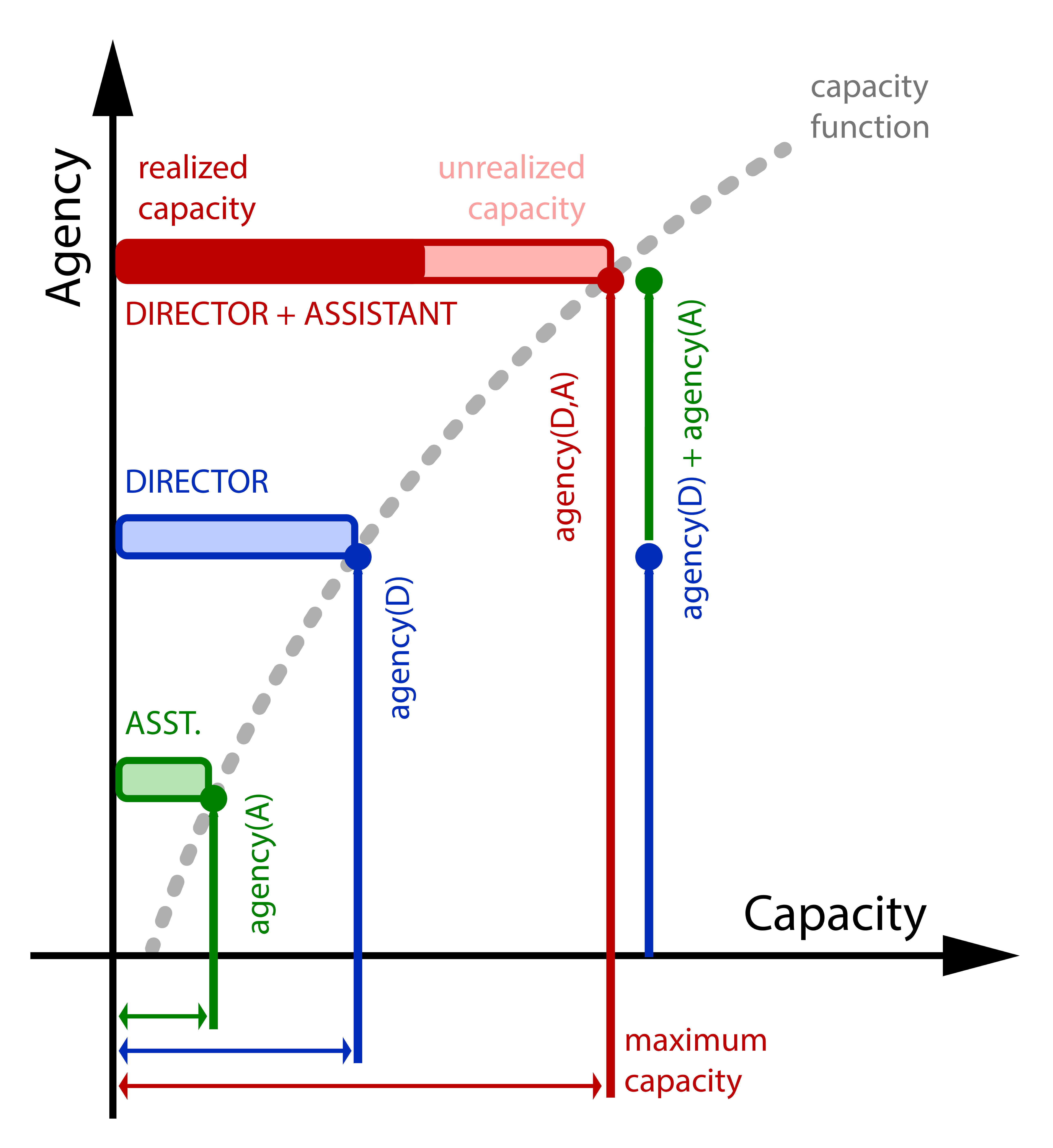}
\includegraphics[height=3.5in]{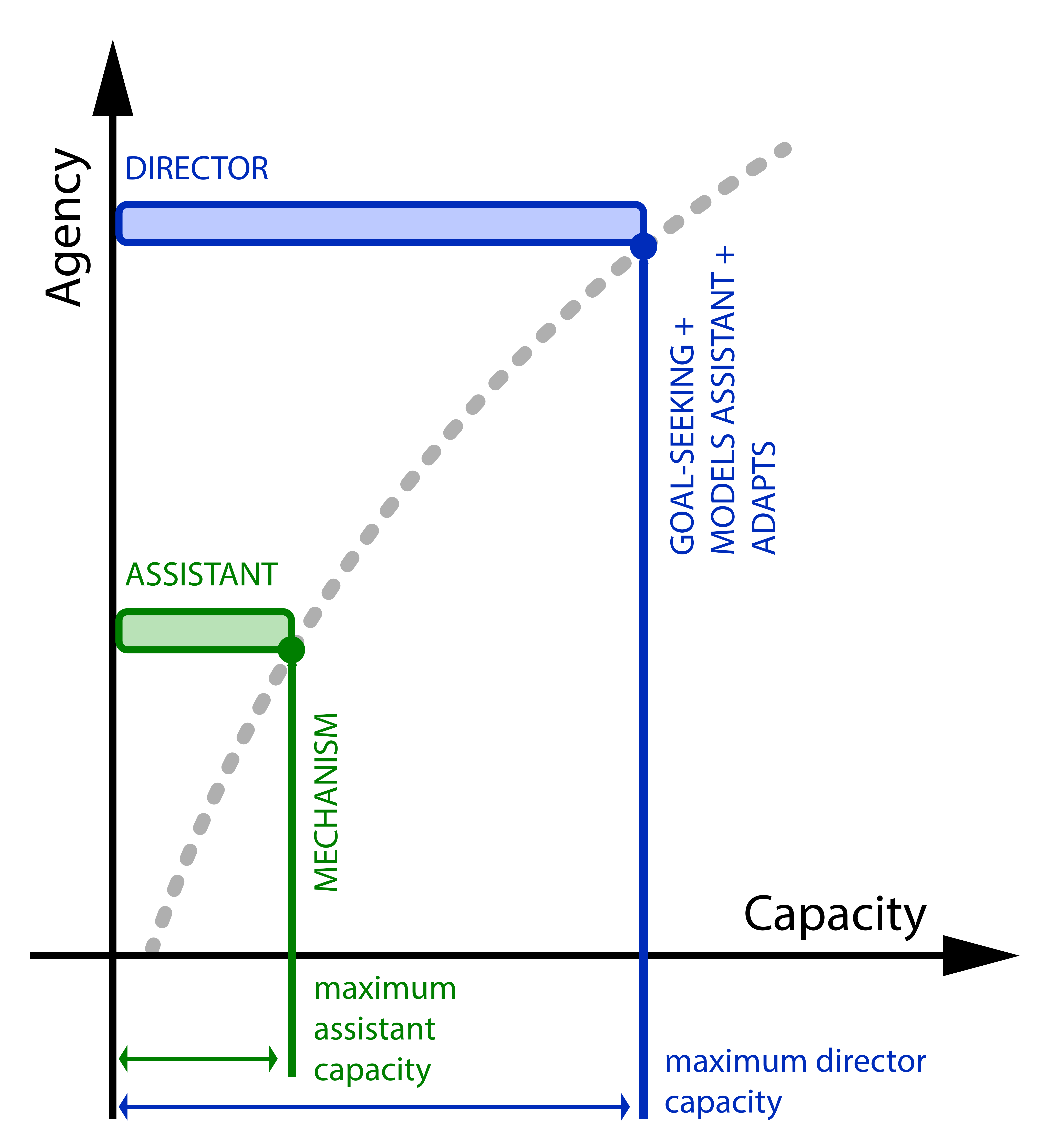}\\
\bf (a) \hspace{2in} (b) \hspace{0.5in}
\caption{Schema presenting the relationship between {\em agency} and {\em maximum capacity} as a point on a curved, measure-specific capacity function (dotted grey line). Shown here are (a) the cumulative nature of the agency of the assistant and director, and (b) the way different attributes of agency combine to increase the maximum capacity of a system or partnership.}
\label{fig:agency}
\end{figure}

With these attributes in mind, we now outline our schema for thinking about different degrees of agency in a prosthetic setting and relating that agency to the combined capacity of a director-assistant partnership. Figure \ref{fig:agency} presents the key parts of the schema in visual form. Capacity and agency in this schema are deliberately presented below in a way that is agnostic to units of measure and the exact attributes of agency, so as to be compatible with and still helpful across multiple definitions of agency.

{\bf Capacity:} We define {\em capacity} as the capability or prowess achieved by the director-assistant partnership. This capacity is defined with respect to one or more measures of performance, with the combination of all such measures approximating the true capability of a partnership. {\em Maximum capacity} is defined as the best possible performance for a given measure that could be achieved by the partnership, shown as a shaded red bar in Fig. \ref{fig:agency}a (e.g., the fastest time to complete a manipulation task or the number of repetitions of a performance task completed; in practice, the maximum capacity may only be able to be approximated using data from a normative subject population.) {\em Realized capacity} is defined as the current ability of the partnership according to a given measure or set of measures (solid red bar in Fig. \ref{fig:agency}a). 

{\bf Agency:} Agency is here visualized as a measurable, continuous quantity that is a summation of contributions from individual attributes of agency. In plain terms, multiple attributes of agency combine together to increase the overall agency of the director (Fig. \ref{fig:agency}b, blue), assistant (Fig. \ref{fig:agency}b, green), or the director/assistant partnership (Fig. \ref{fig:agency}a, red). For example, the agency of a system determines its placement on the vertical axis in Fig. \ref{fig:agency}; the summation of the agency of the director and assistant (blue and green vertical lines in Fig. \ref{fig:agency}a) defines the agency of the partnership. 

{\bf Capacity function:} Agency can be mapped to capacity by way of a measure-specific curve that we here call a {\em capacity function}: the relationship between agency and capacity that is defined by the physical and informational priorities of systems and environments (and potentially unknown to a designer). By finding the point on a capacity function corresponding to a given level of agency, we can visualize the maximum capacity of partnership. For example, in Fig. \ref{fig:agency}b, when agency is mapped upwards to a capacity function's curve (grey dotted line) a system that is a pure mechanism (here the assistant) will result in less maximum capacity than a system that has goals, models its assistant as a mechanism, and adapts to changing circumstances (here the director). The combination of such a director and assistant would result in greater maximum capacity than the sum of the two individual systems (as plotted on an arbitrary capacity curve in Fig. \ref{fig:agency}a).

\begin{figure}[t]
\centering
\includegraphics[height=2.5in]{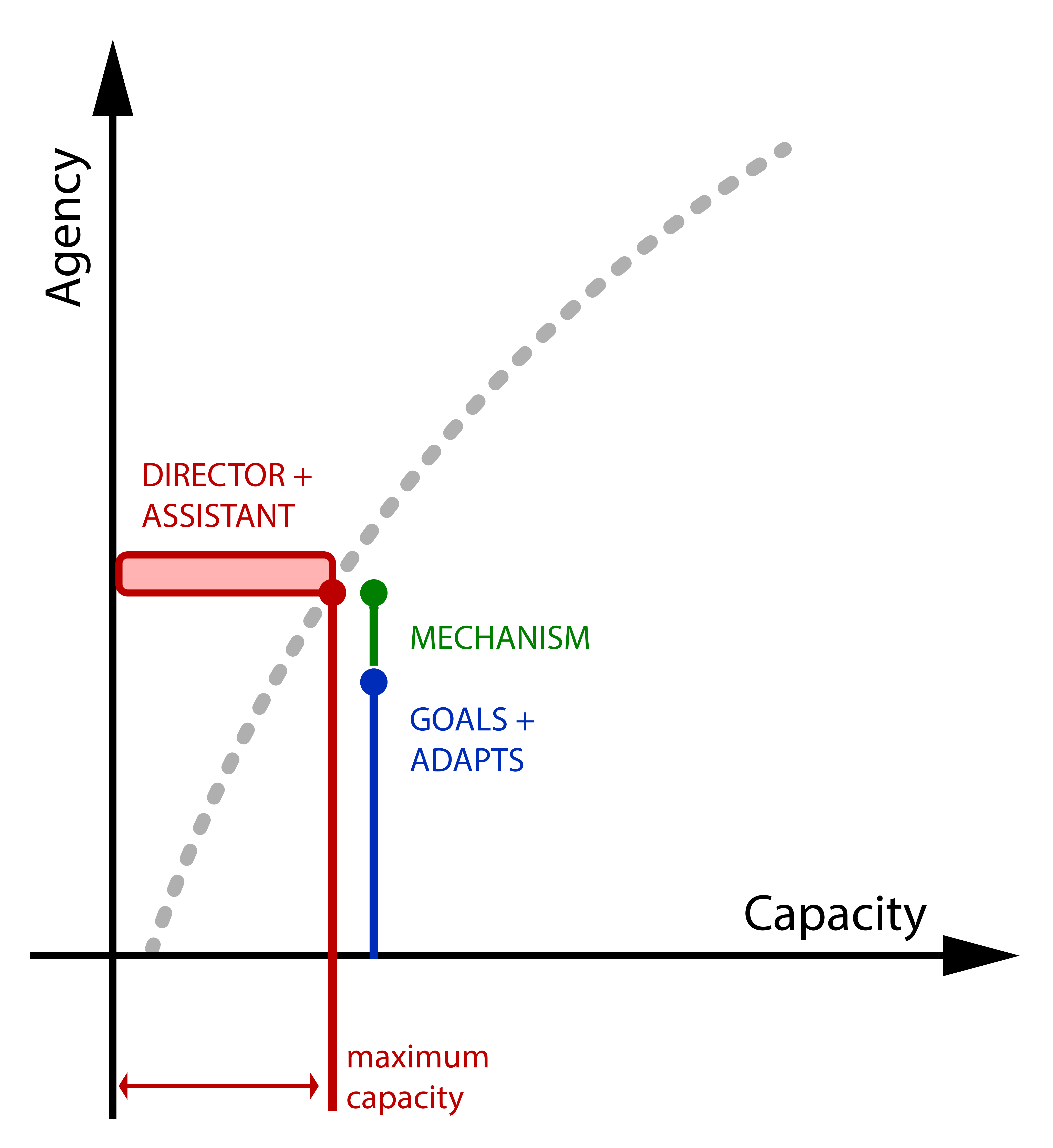}
\includegraphics[height=2.5in]{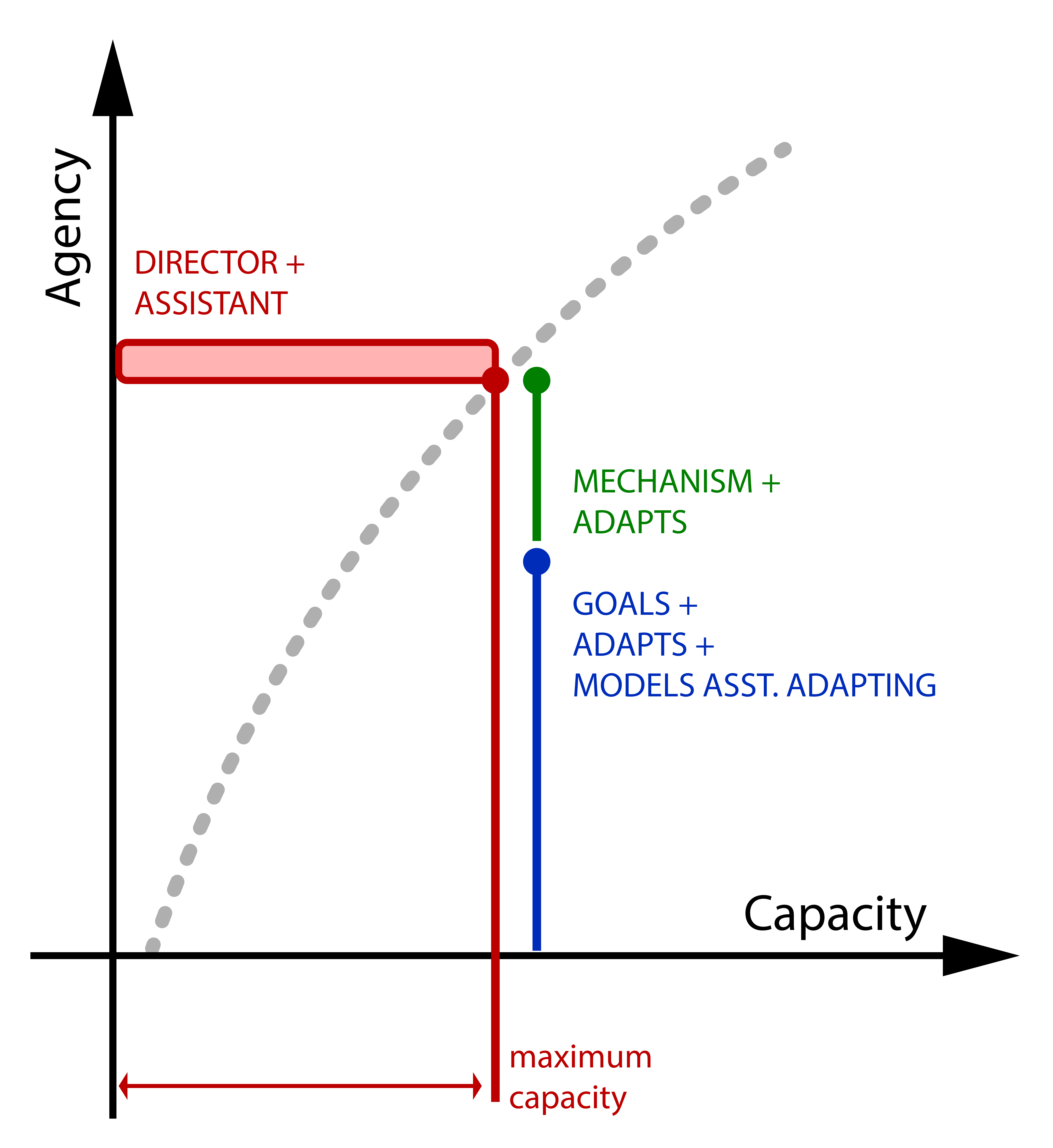}
\includegraphics[height=2.5in]{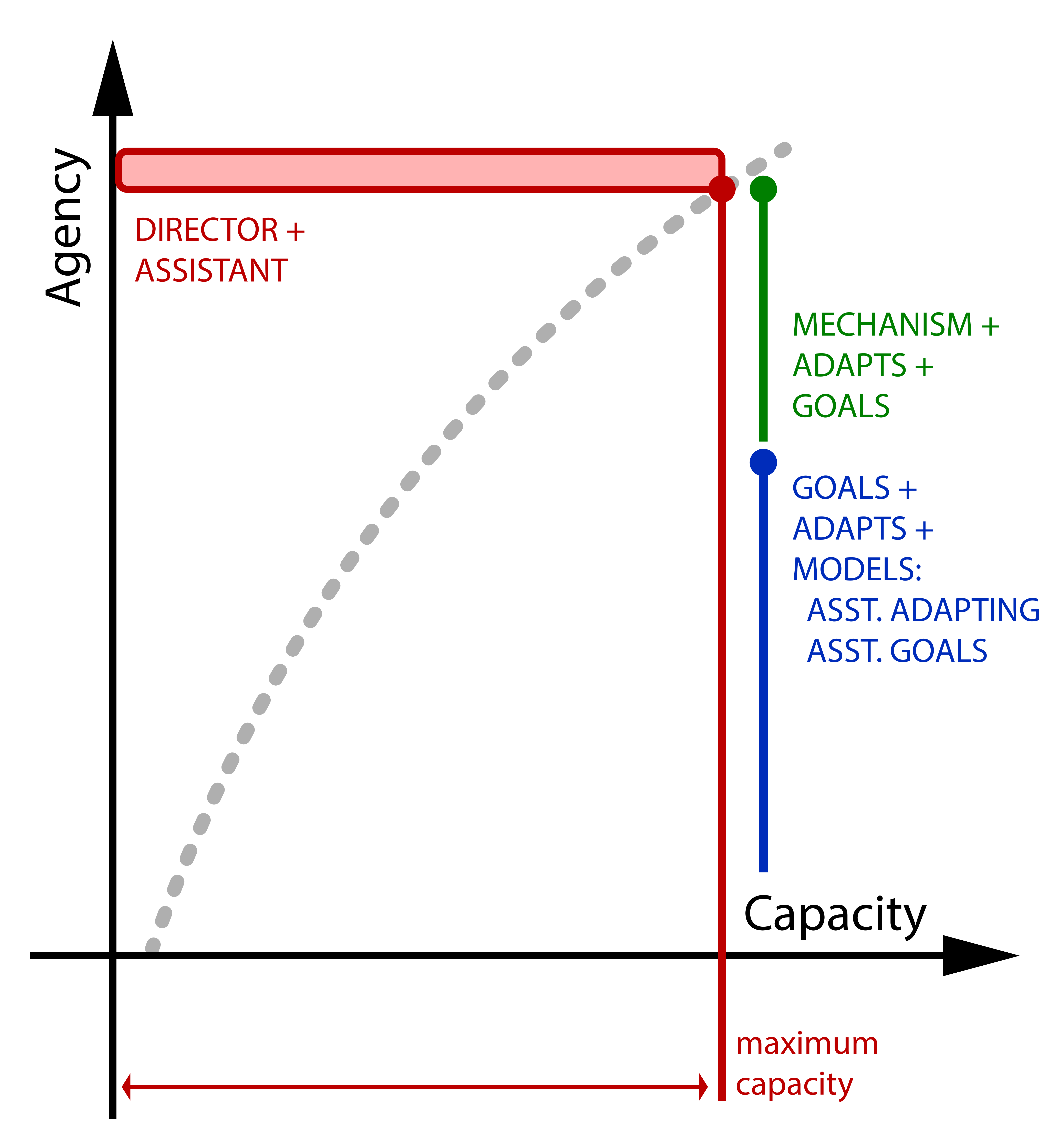}
\bf (a) \hspace{1.5in} (b) \hspace{1.5in} (c)
\caption{Combined attributes of agency and the resulting capacity. a) The standard setting for prosthetic control, where a human director with goals utilizes a fixed mechanism;  b) the case demonstrated by Edwards \etal\  (2015), and in commercially deployed pattern recognition, wherein the human interacts with a prosthetic system that they know acquires, adapts, and uses predictions (knowledge) in control; c) the case where the assistant has knowledge-supported goals and the director views the assistant as having knowledge-supported goals---for example, reward-based training of a myoelectric controller by Pilarski \etal\  (2011) and Mathewson and Pilarski (2016). (These cases are each described in detail in Sec.\ 6.)}
\label{fig:combs}
\end{figure}

To further clarify how capacity varies with agency in this schema, Fig. \ref{fig:combs} depicts three example combinations for the director and assistant that can be readily identified from the contemporary literature.
In these examples, and those that follow, we assume that a human director has a set of goals that relate to their task, needs, and environmental setting. Defining the goal of an assistant is not always straightforward, but one possible and immediate goal for the assistant is gaining approval from the director. Approval may be communicated to the assistant via any of the normal communication channels between the two partners, or through a privileged channel dedicated to reward. Like an assistant in the corporate sense, a goal-seeking prosthetic assistant could strive to maximize director approval, but would also have its own goals that may be overridden by directions from the director. For example, the assistant may have the goal to protect itself at the onset (to prevent its motors from overheating during use or its battery from running dead), but in order to secure the director's approval, have the capacity to align its behaviour over time to the director's goals as they become clear to the assistant. 
For the cooperative director-assistant relationship, we expect that each will be able to best support the goal-driven behaviour of the other only when they can fully model the other and understand the other's goals. 

How the goals of the director and assistant can come to align in a general sense is an interesting problem, one not far removed from team formation in human-human or human-animal partnerships (e.g., Fishman 2003). It is very natural to think that such alignment can occur during normal sensorimotor interactions between agents (Pezzuolo \etal\ 2011 \& 2013, Sebanz \etal\ 2006 \& 2009). To examine the process by which such alignment might occur during human-prosthesis interaction, we now introduce the idea of communicative capital.

\section{Communicative Capital}

As depicted in Figs. \ref{fig:agency}a and \ref{fig:progression}, agency contributed by a director and an assistant defines the maximum capacity of a director-assistant partnership. {\em Communicative capital} is the communication infrastructure that has been built or otherwise acquired by both sides of a partnership to facilitate the pursuit of this maximum capacity.  Communicative capital is what enables the partnership to realize capacity towards that maximum capacity---i.e., to increase the portion of the red bar that is solid and not shaded in Fig. \ref{fig:agency}a. More precisely, communicative capital is defined here as a set of communication resources that are built up through ongoing interactions between a director and an assistant. 

One way of thinking about progressively better interactions between an assistant and a director is as investing and accumulating communicative capital. Like investing in capital in the economic sense, communicative capital requires cost in terms of effort to establish and maintain (c.f, the signaling costs described by Pezzulo \etal\ (2013)). This cost may be incurred during the normal interactions of a partnership, or, in many cases, through special effort that is orthogonal to the explicit goals of the partnership. Users of prosthetics devices are required to learn about the use of their prosthesis before they take it home for use in activities of daily living; they build up an understanding of the mechanisms of the device. In more advanced examples like pattern recognition, teaching one or both sides of an partnership a language for communication (e.g., a series of commands to a prosthesis phrased in terms of patterns of myoelectric signals) may delay communication during teaching, but increase the efficiency of communication later on. A simple case of this is when one stops a conversation to define a new term or command so that subsequent communication and control can be more clear and efficient. This term and the meaning it conveys to both sides of the partnership is an example of communicative capital. The use of communicative capital to realize capacity may also have an ongoing cost, such as when a director makes their actions intentionally more clear or predictable so as to better inform their assistant (Pezzulo \etal\ 2013).

Another way to view communicative capital is a process of compression and decompression. The sender takes an action or speaks a word which acts as a token to represent significant amounts of other information such as associations, memories, relationships, implication, and intentions. Examples include the information communicated by hesitation movements when two agents both reach for the same object (Moon \etal\ 2013), or the precisely shaped kinematics of participants in a task where they must coordinate their actions to achieve a goal (Candidi \etal\ 2015). The sender may not even be aware of all that is represented. When a token is received by the receiver a reverse mapping occurs, a decompression, which attempts to recover the associated information. This reverse mapping may not be complete, may at times be incorrect, and may even add additional information.

In any interaction, we expect systems with differing levels of agency to form communicative capital at different rates, in different quantities, and of different complexity. To begin to form communicative capital, at least one system needs to be above the level of a non-adaptive mechanism (i.e., systems must be able to adapt). Further, we expect the greatest opportunities for communicative capital and a fruitful progression of interactions when both the director and the assistant exhibit the highest possible degrees of agency. We now discuss how communicative capital may be built and used to progressively realize more capacity in prosthetic director-assistant partnerships.

\section{Building Communicative Capital through Interaction}

\begin{figure}[t]
\centering
\includegraphics[height=2.7in]{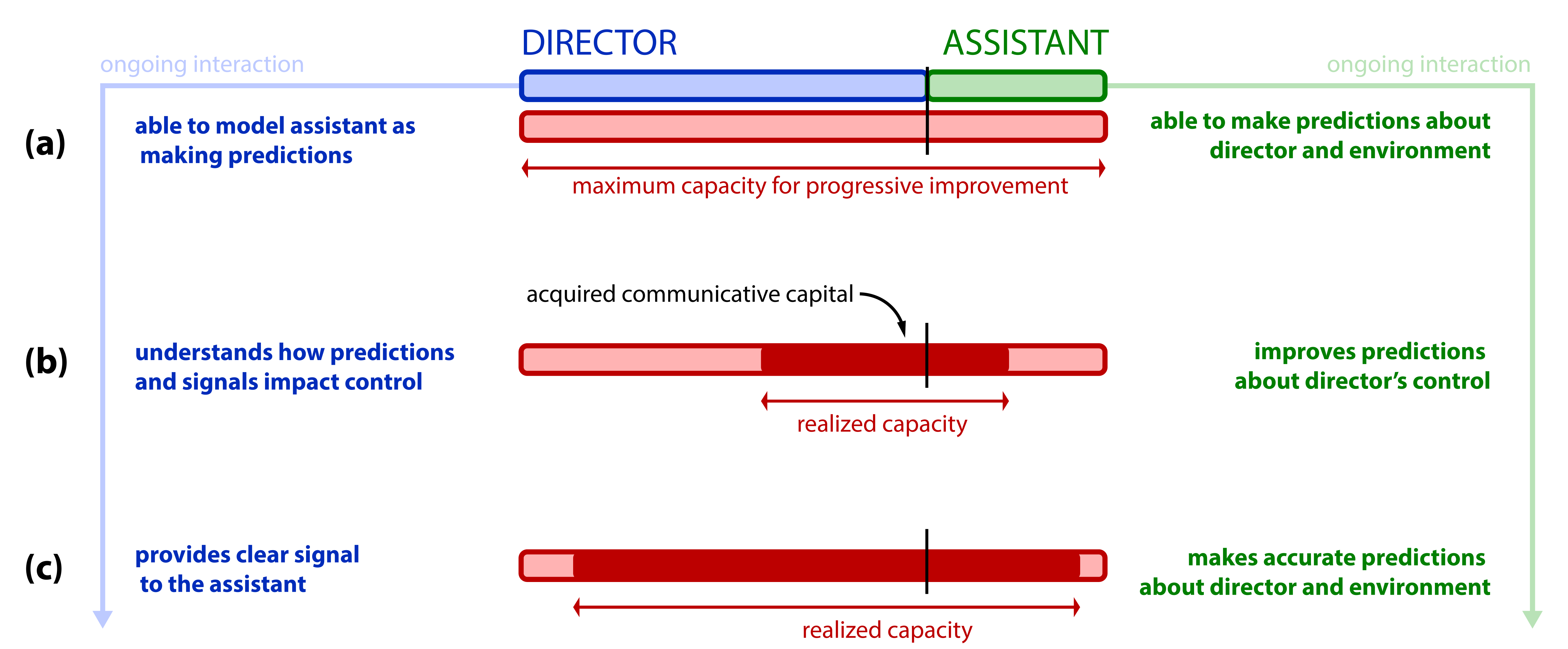}
\caption{An example progression wherein the combined ability of the director and assistant approaches its full capacity via ongoing human-machine interaction. Over time and through better modeling of each others' behaviour and the environment, the systems build up {\em communicative capital} that can be used to achieve their goals. One natural type of communicative capital is knowledge in the form of predictions.}
\label{fig:progression}
\end{figure}

So far we have considered the setting in which a communication channel is opened up between the director and the assistant. This channel could be either unidirectional or bidirectional. It should be expected that for interactions between two goal-seeking agents, communication will improve the lot of both the director and the assistant. If the receiver's goals are not furthered by the information received, then it may ignore the received information. If the sender's goals are not furthered by what the receiver does with the information, then the sender will not send it. The sender can send many possible things, and can therefore choose how to balance the cost of sending information with the expected outcomes for the agent and the partnership (Pezzulo \etal\ 2013). It follows that the sender should vary its communication to find the information to send that results in both systems improving with respect to their goals. The variation of communication could be independent, or guided by other parties---e.g., the work of clinical staff during training a patient for prosthesis use, or an instructor helping someone collaborate with an assistive animal (Pfaffenberger \etal\ 1976). 

In effect, the processes of building communicative capital toward the attainment of goals is about the specification and identification of things the sender cares about, as in ``when I say or do {\em this} it means {\em this}''. There can then be a natural progression in the interaction as the two sides get to know each other better---for example, the simple progression shown in Fig. \ref{fig:progression}. As depicted in Fig. \ref{fig:progression}, acquired predictions and anticipations represent one powerful form of communicative capital. As in the case of an executive assistant to the director of a company, good assistance often relies on the capacity of the assistant to forecast the director's needs (what they want and when they want it) such that it is possible for the partnership to achieve tasks with minimal effort and with minimal communication on the part of the director. This viewpoint is very compatible with contemporary perspectives on human-human motor coordination (Candidi \etal\ 2015) and with prosthetic control approaches like pattern recognition (Micera \etal\ 2010, Scheme and Englehart 2011), as will be discussed below.

In the following sections, we use the idea of communicative capital and the agency-capacity schema defined in Sec. 4 to examine three broad classes of experimental work from our group and others where prosthesis control has been improved by ongoing interactions between the device and the user. These classes include human interactions with clinically prevalent {\em mechanisms} for prosthesis control, with {\em adaptive mechanisms} like pattern recognition, and with {\em goal-seeking prosthetic agents}.

\subsection{Mechanisms: Conventional Control}

\begin{figure}[t]
\centering
\includegraphics[height=2.7in]{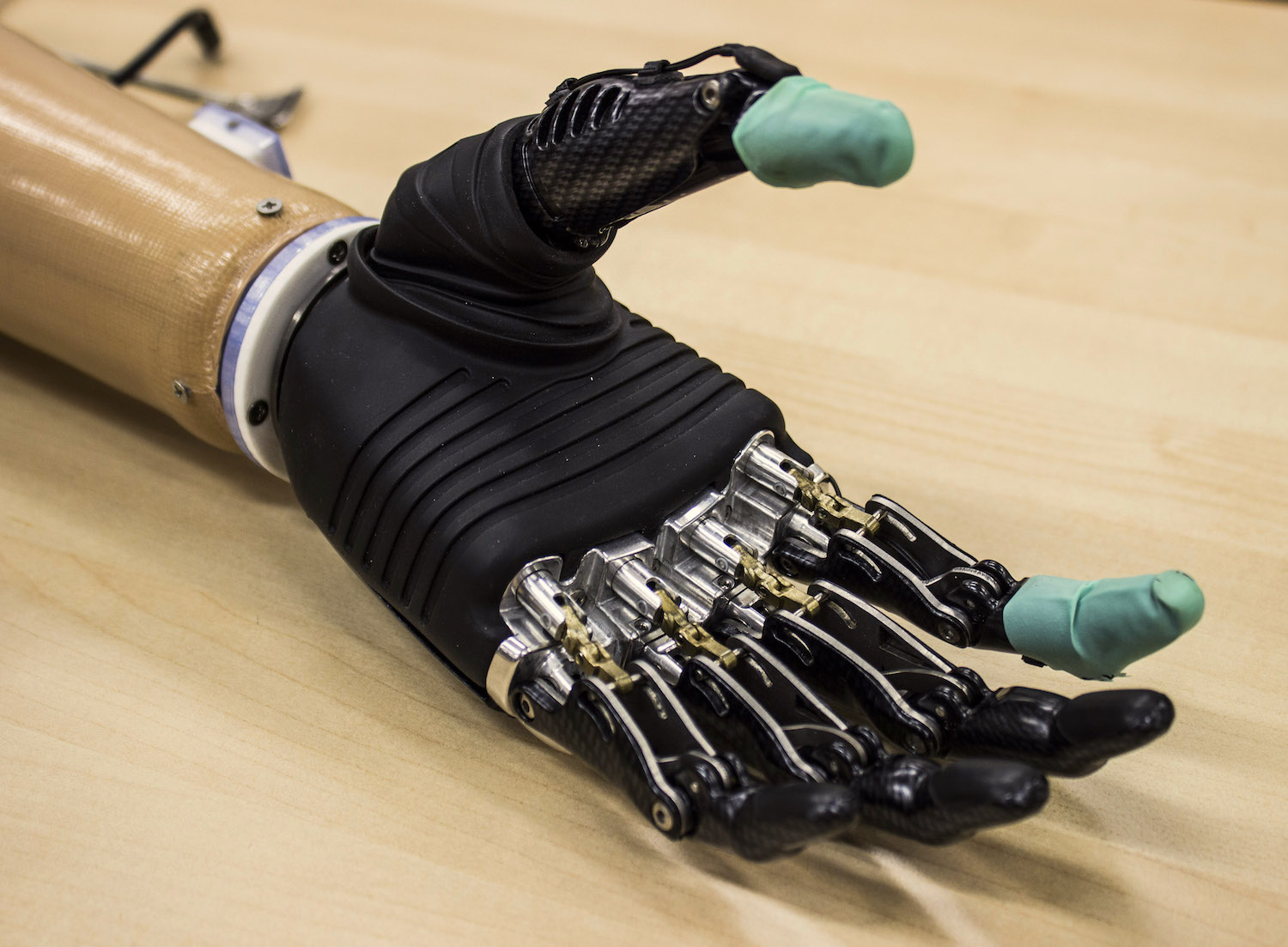}
\includegraphics[height=2.7in]{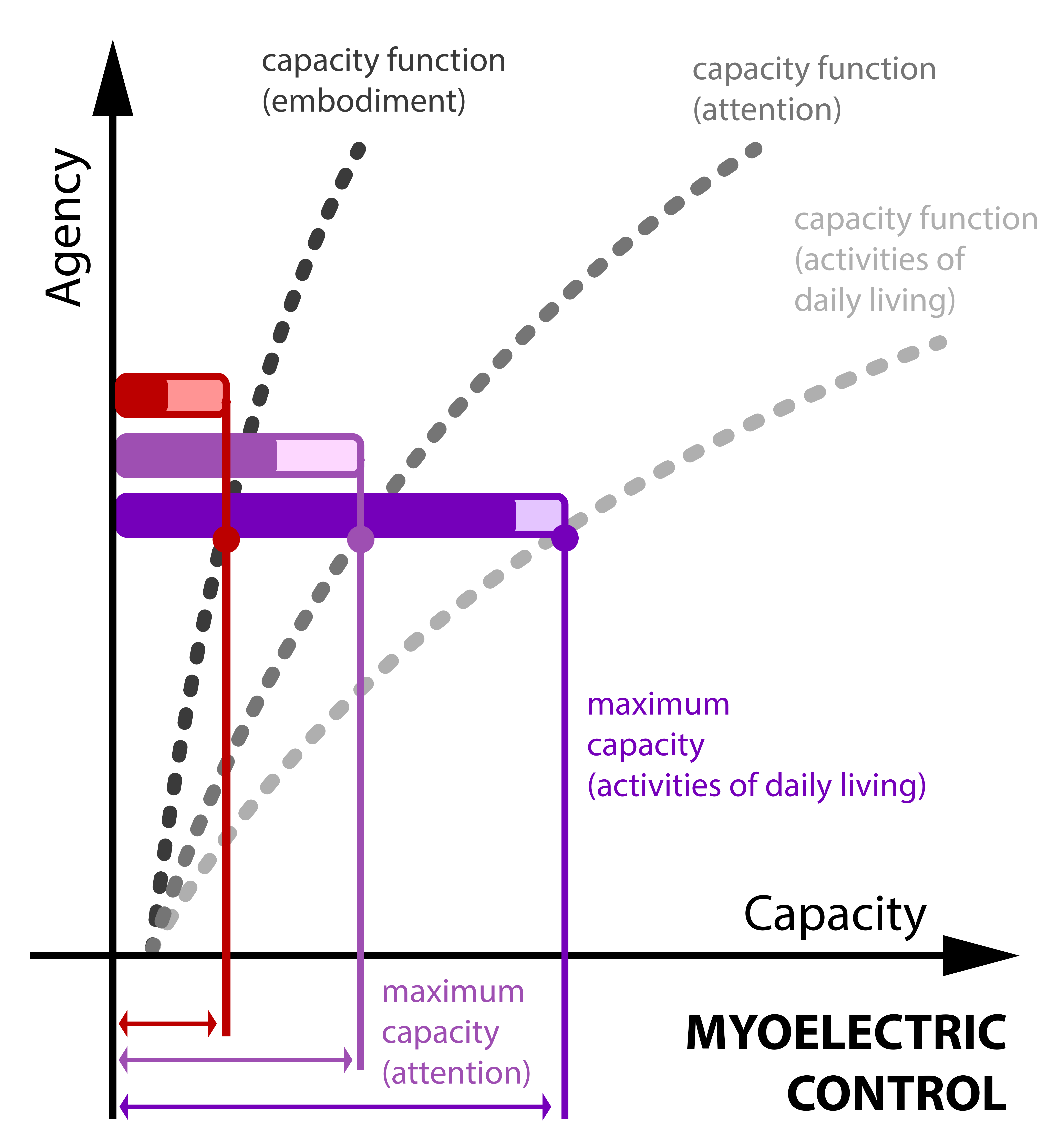}\\
\bf (a) \hfil (b)
\caption{A common setting: (a) human interactions with a non-adaptive mechanism in the form of a myoelectric prosthesis. (b) Interactions are examined using the agency schema from Sec.\ 4, with selected capacity functions relating to daily life activities, required intentional load, and prosthetic embodiment.}
\label{fig:myocontrol}
\end{figure}

Conventional control interactions between someone with an amputation and their powered prosthesis represent the dominant case of a goal-seeking agent (the director) interacting with a mechanistic assistant (Fig.\ \ref{fig:combs}a). In this setting, the prosthetic assistant interprets electromyographic (EMG) signals from the director's remaining muscles and maps these signals in a fixed way to control commands for one or more robotic actuators, an approach known as {\em myoelectric control} (Parker \etal\ 2006, Micera \etal\ 2010). The only agent capable of progressive change is the director, who learns to better use the myoelectric control mechanism to achieve their goals. The capacity of the complete system is therefore a function of the director's ability to learn, improve, and adapt, along with a fixed (but potentially significant) contribution from the nature of the assistant. A comparable analogy is an elite athlete adapting to their sporting equipment. We know from clinical experience that training is a large part of successful myoelectric control by people with amputations.

Figure\ \ref{fig:myocontrol} shows three example capacity curves for the setting of a mechanistic assistant: the ability to complete activities of daily living, the attention a director needs to devote to the assistant (e.g., eye contact on the terminal device), and embodiment (the way the assistant is integrated into the director's body image or sense of self). With training, it is well known that a user can progress towards the maximum capacity or performance as dictated by their contributions as a goal-seeking agent and the contributions due to the fixed nature of the mechanism of the assistant and its control system. It is also well known that capacity may be realized at different rates for different measures, e.g., someone may be very fast to learn new activities, but take more time to reduce their direct attention to the device or begin to feel like it is part of their body. Common clinical examples of communicative capital built by the director of a myoelectric prosthesis include the director interpreting vibrations and sounds of the robot's motors to inform their control. Human learning allows the director to optimize their sensorimotor interactions with a powered or unpowered prosthetic assistant.

\subsection{Adaptation: Predictively Enhanced Control}

Some of the most appealing and most prevalent examples of communicative capital in the prosthetic setting stem from adaptive control paradigms---specifically, machine learning based prosthetic controllers. There are multiple examples of director-assistant interactions in prosthetics where the director views the assistant as adapting (in this case as making predictions or control forecasts) and where the assistant acquires knowledge about the director to better execute the director's intention (the progression depicted in Fig.\ \ref{fig:progression}) (Castellini \etal\ 2014, Edwards \etal\ 2015 \& 2016, Pilarski \etal\ 2013a). 

\begin{figure}[t]
\centering
\includegraphics[height=2.7in]{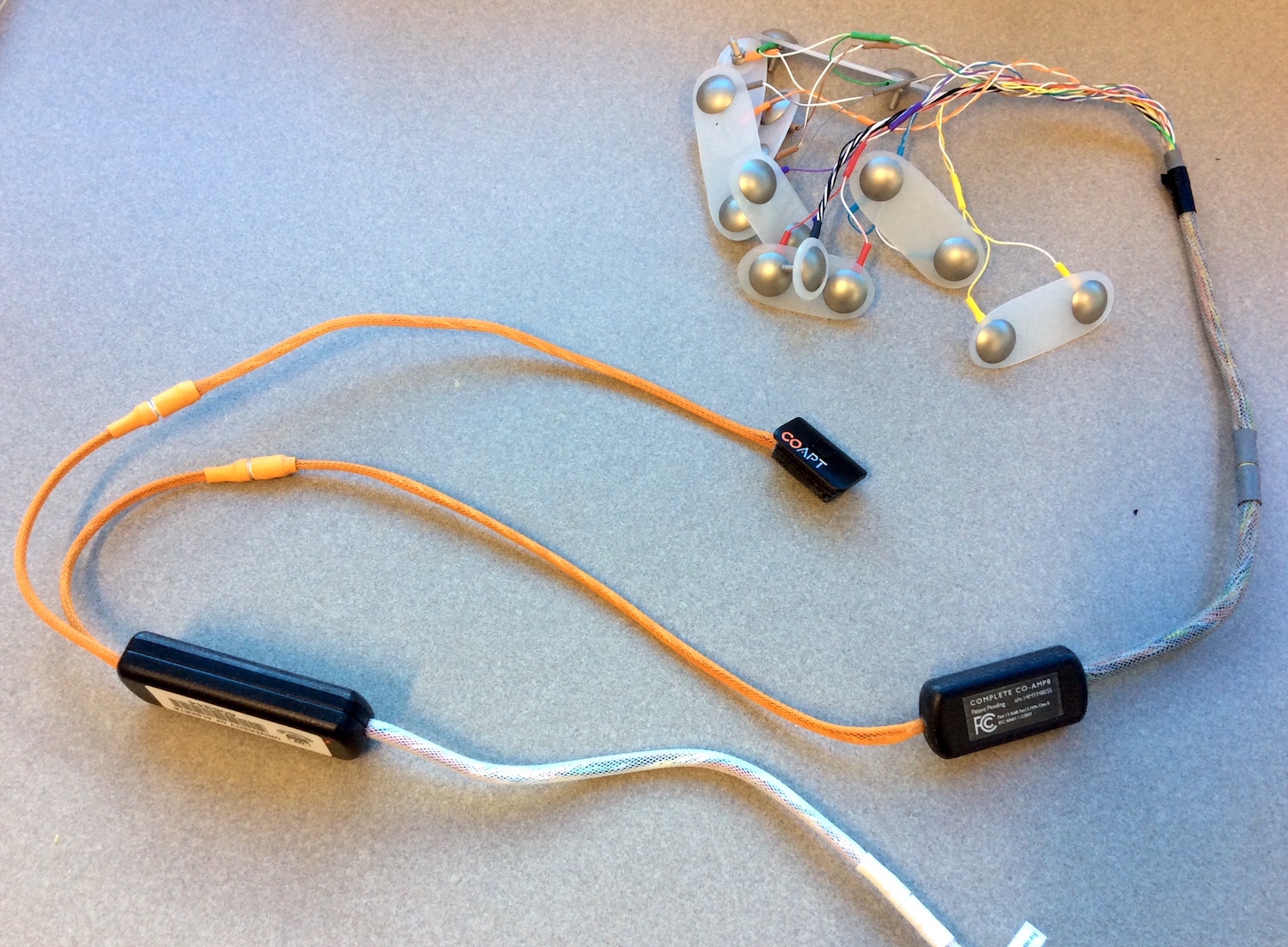}
\includegraphics[height=2.7in]{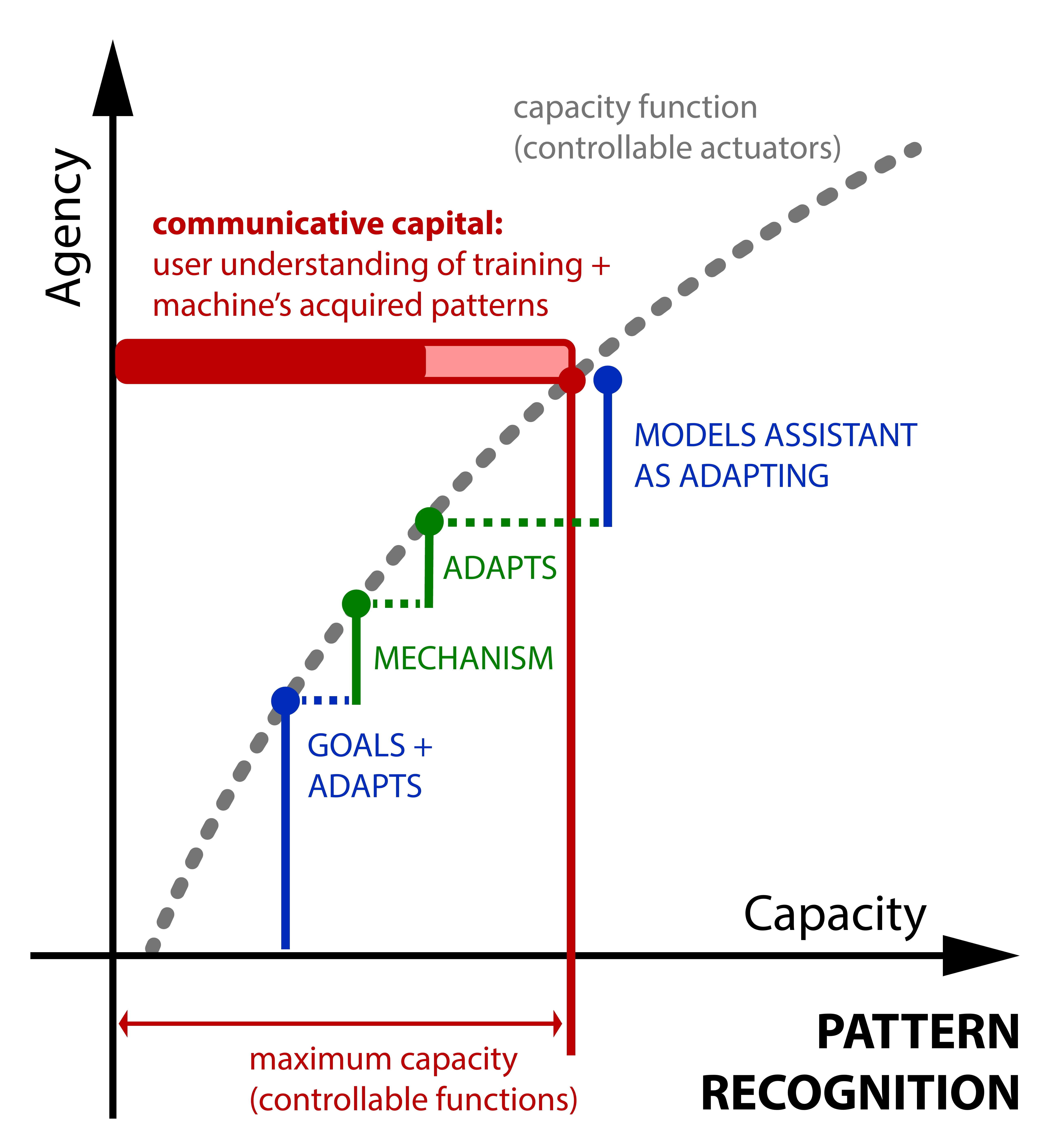}\\
\bf (a) \hfil (b)
\caption{Example of an adaptive prosthetic assistant: (a) pattern recognition is a commercially available prosthetic control system based on machine learning, wherein the director trains the assistant to interpret myoelectric signals so as move and coordinate the different joints of a prosthesis. (b) Using the agency schema from Sec.\ 4, the use of pattern recognition allows the assistant to adapt, and in doing so, the director begins to think of the assistant as a system that adapts; the maximum and realized capacity of a director-assistant partnership, in concrete terms of the number of accessible functions or degrees of control, is increased through the use of pattern recognition. Communicative capital is formed via the assistant's improved predictions about the user's control intent, and the director's understanding of how best to train and provide signals to the assistant.}
\label{fig:patternrec}
\end{figure}

A first example is {\em commercial pattern recognition} (Fig. \ref{fig:patternrec}), wherein the director is able to engage a training phase to inform the prosthetic assistant about the right motions to perform in response to complex patterns of myoelectric activity recorded from the director's body (Castellini \etal\ 2014, Scheme and Englehart 2011). The use of pattern recognition can provide users with more intuitive control of their prosthesis (Castellini \etal\ 2014). The director becomes more skilled at providing clear training commands, in part because of their knowledge that the assistant is learning from their demonstrations (Fig. \ref{fig:patternrec}b). The result is improved capacity in terms of the number of controllable functions that can be accessed by a director (the capacity function in Fig. \ref{fig:patternrec}b), in practice far exceeding the number of available degrees of control in conventional myoelectric control.

\begin{figure}[t]
\centering
\includegraphics[height=2.7in]{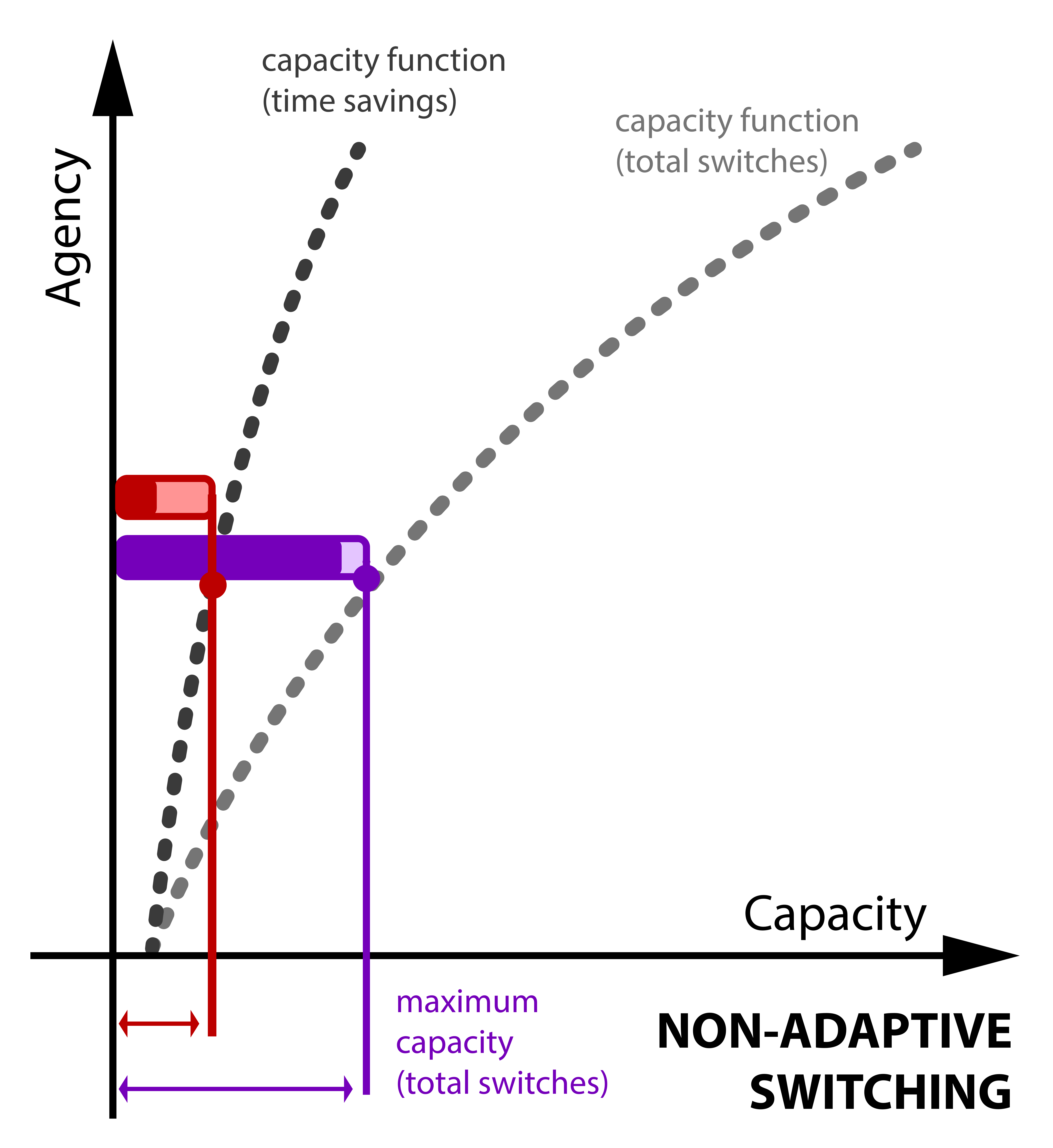} \hspace{1em}
\includegraphics[height=2.7in]{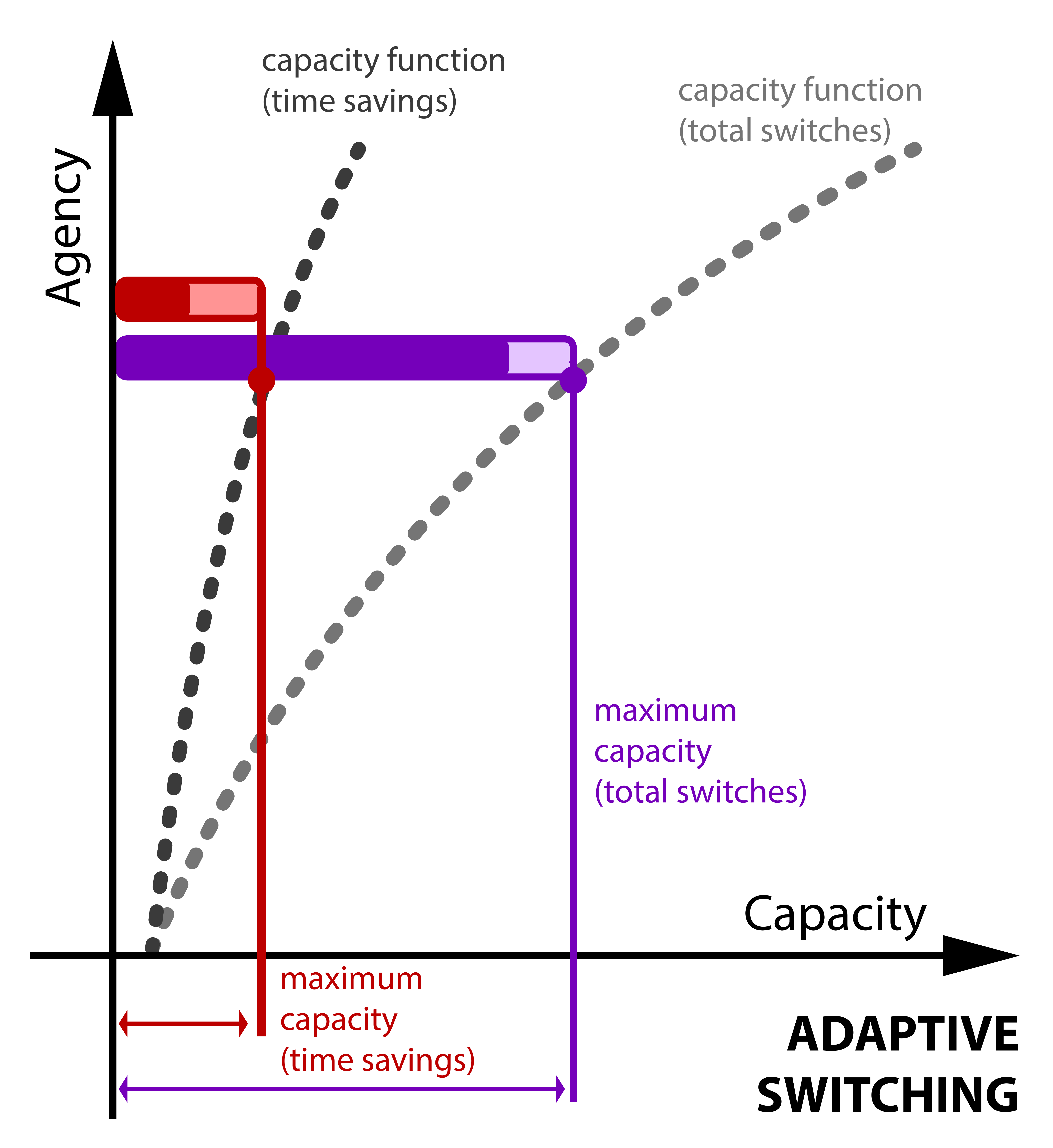}\\
\bf (a) \hfil (b)\\
\vspace{0.6em}
\includegraphics[height=2.5in]{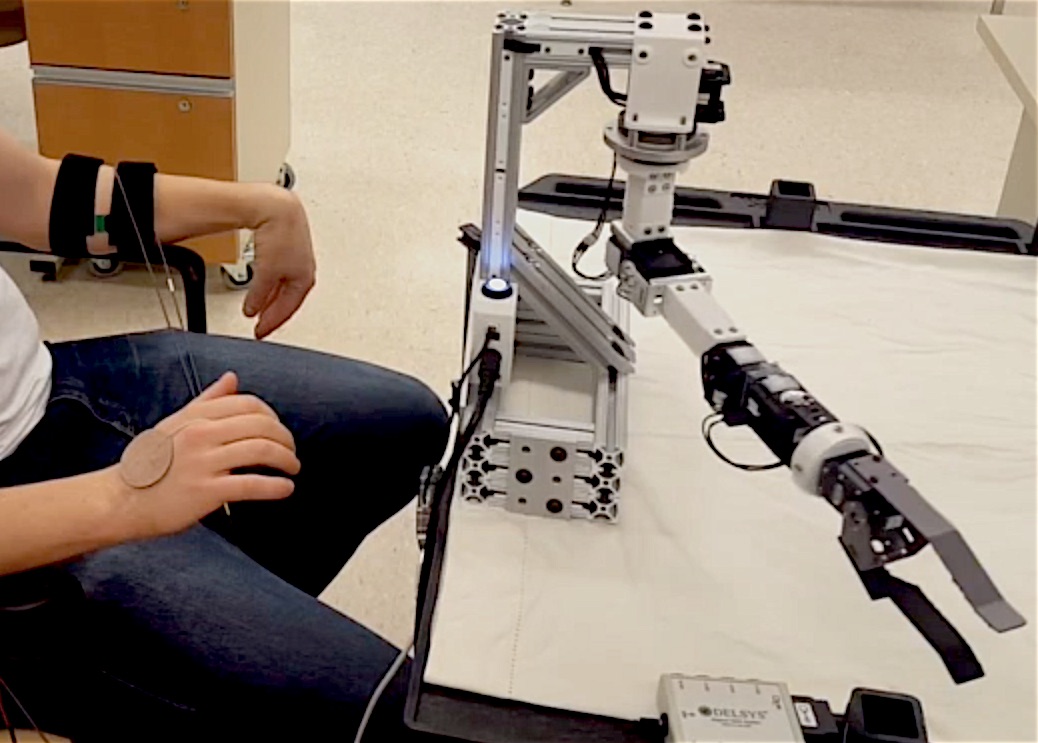}\\
(c)
\caption{A second example of an adaptive prosthetic assistant: adaptive switching allows a prosthetic assistant to learn the way a director uses the many functions of a prosthesis, and thereby streamline the director's control interactions. (a) With a non-adaptive assistant, the  maximum capacity of the partnership depends on a director's ability to adapt to the assistant, as in conventional control, and improvement is limited in terms of the time to complete a task and the total number of manual interactions required of the director (shown capacity functions). (b) With adaptive switching, the assistant has been shown  to adapt to the director, decreasing the time to complete tasks and significantly reducing the number of required manual interactions during (c) interactions between a amputee and non-amputee human directors and a desk mounted robot arm  (Edwards \etal\ 2015 \& 2016). Communicative capital is formed via the assistant's improved predictions about the user's switching behaviour, and the director's understanding of how the assistant will suggest prosthetic modes or functions in different situations. As in Fig. \ref{fig:patternrec}, the cumulative increase in agency of the partnership over the mechanistic case is a direct result of the adaptive nature of the assistant, and the ability of the director to think of the assistant as adapting.}
\label{fig:adaptive}
\end{figure}

\begin{figure}[t]
\centering
\includegraphics[width=0.55\linewidth]{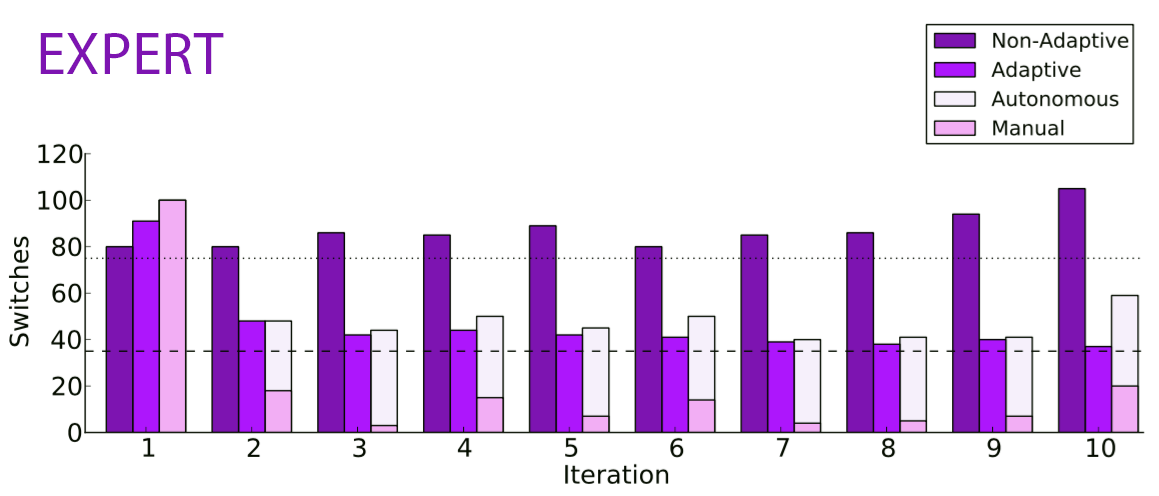}\\(a)\\
\includegraphics[width=0.55\linewidth]{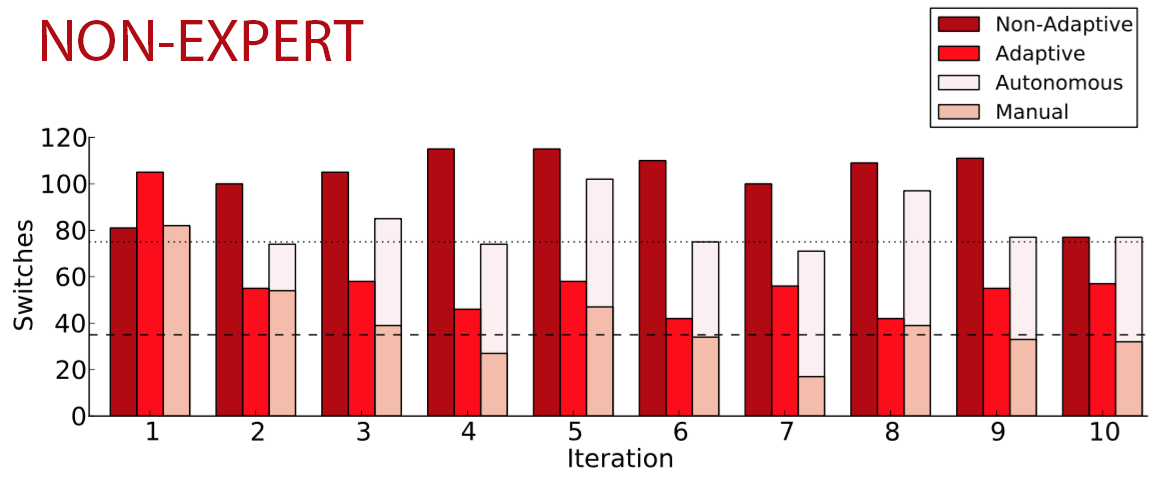}\\(b)\\
\includegraphics[width=0.55\linewidth]{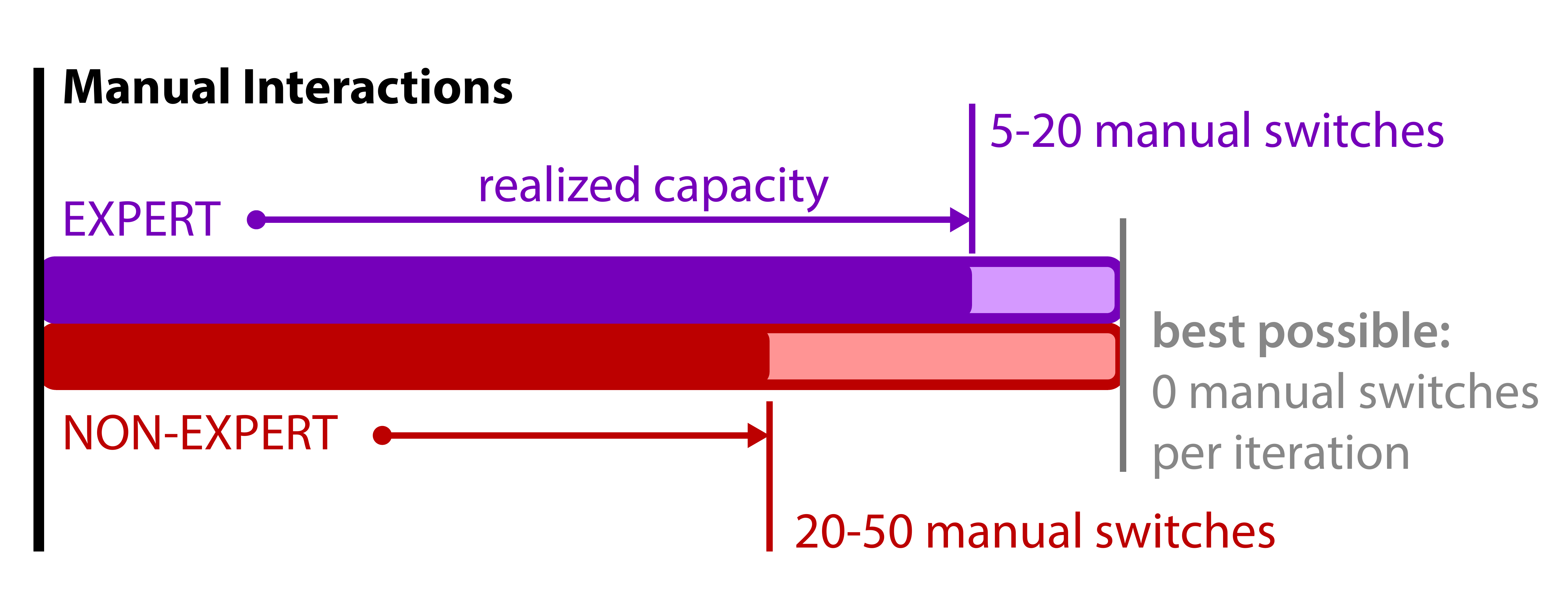}\\(c)\\
\includegraphics[width=0.55\linewidth]{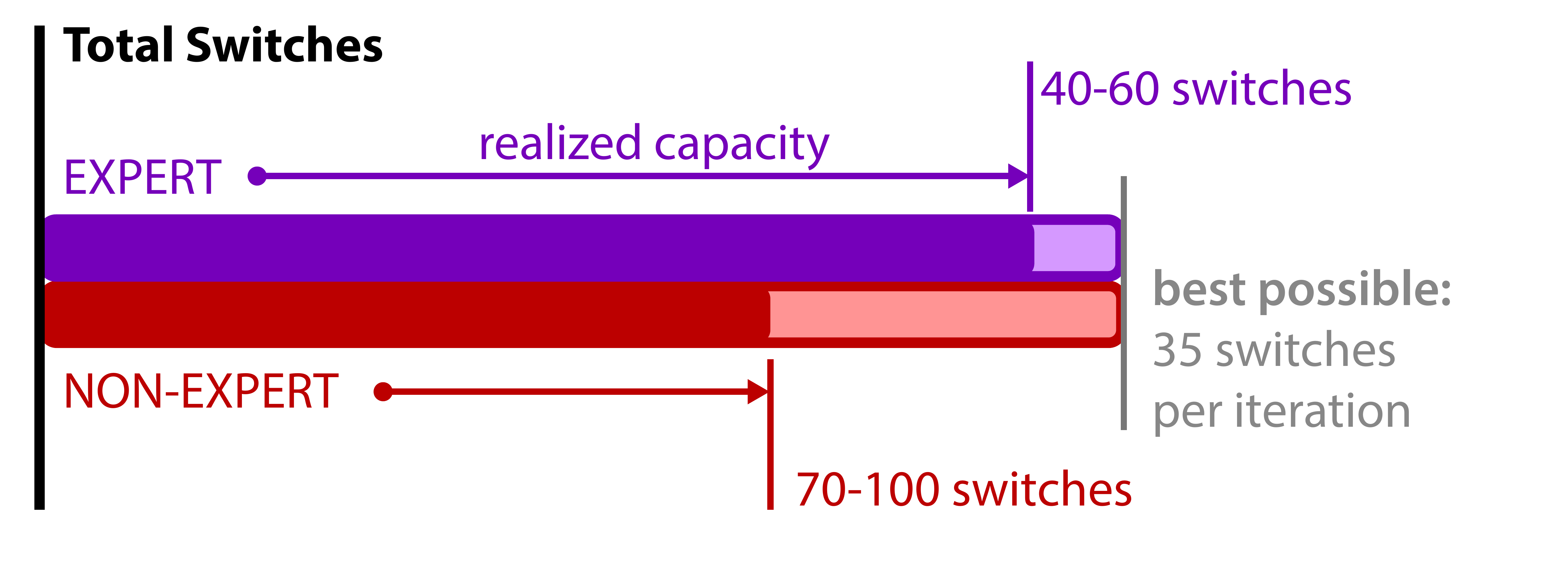}\\(d)
\caption{A comparison of realized and maximum capacity for autonomous and adaptive switching, with data plots showing (a) expert and (b) non-expert directors (plots adapted from Edwards 2016). As seen for capacity functions relating to (c) the number of manual interactions and (d) the total number of switches required to complete a myoelectric control task, expert directors were able to realize more capacity than non-expert directors.}
\label{fig:autoswitch}
\end{figure}

A second example is {\em adaptive and autonomous switching} (Edwards \etal, 2015, 2016; Edwards 2016), wherein the assistant learns and makes ongoing predictions about how a director will switch between the many functions of a prosthetic device (Fig. \ref{fig:adaptive}). In adaptive switching, the director's ability to quickly execute tasks is improved by using switching suggestions made by the assistant; at the same time, the assistant improves its suggestions based on ongoing observations about the director's actions and preferences. The adaptive nature of the assistant and the increased agency of the director to model the assistant can be seen to lead to increased maximum and realized capacity in terms of reducing both task time and total switches needed by a human user to complete a task (Fig. \ref{fig:adaptive}a,b). In autonomous switching, the assistant makes predictions and uses these predictions to automatically switch between the functions of the prosthetic device (Edwards 2016, and Fig. \ref{fig:autoswitch}a,b). As noted above, predictions form a very natural and readily acquirable form of communicative capital that can be built up by a machine learning assistant during its interactions with a director and the world. Further, we can view the acquisition of predictions as happening by way of the same channel as the director's actions, analogous to the action-channel-based signaling examined by Pezzulo \etal\ (2103).

Observations from both adaptive and autonomous switching suggest that the director begins to model the assistant as an agent that makes predictions (Edwards 2016). As subjects became more familiar, both with their execution of a task and with the role of machine learning as it adapted to a task, they reported greater trust in the autonomy of the assistant. In the experiments of Edwards (2016) and Edwards \etal\  (2015), certain regions of task spaces were observed where the learning system performed with close to 100\% prediction accuracy; in these regions, it was clear from subjects' behaviour that they felt less need to monitor the prosthetic arm (e.g., the reduced number of manual switches shown in Fig. \ref{fig:autoswitch}). In the autonomous switching experiments of Edwards \etal\ (2016), users began to anticipate autonomous switches, often moving the next prosthetic function prior to hearing a cue alerting them to the machine's automatic switching behaviour. The realization of  capacity in terms of reduced manual switching, and the predictive capital on both sides of the interface that supports it, is  especially evident in users who have extensive prior experience operating adaptive prosthetic devices, and has important implications in human-prosthesis shared control (Fig. \ref{fig:autoswitch}c,d). Users who had a greater understanding of the prosthetic learning system tended to perform actions that benefited learning, allowing the prosthetic arm to build up expectations about their behaviour more swiftly.

A third example is the work of Sherstan \etal\ (2015). In this work both a user and a software agent share control of the movement of a robotic arm. The user is able to control a single joint of the arm at a time and must switch between joints as needed in order to complete a task. The software agent observes the user's behavior and learns to predict the expected joint angles of the robot arm. These predictions are then used by the agent to move the arm in collaboration with the user's own actions. This is a limited form of learned communicative capital. Here the agent is learning to interpret the intent of the user based on their action. This example is akin to two people, a director and an assistant, moving a large piece of furniture together (c.f., Sebanz \etal\ 2006). When the assistant observes the director begin to move left the assistant anticipates that the director wants to move into the hall and moves their end of the furniture to the right to accommodate the turn. Further, more indirect cues might be interpreted, such as the director preemptively turning their head to the left in anticipation of the hallway.

As a final example of adaptive assistive technology outside the prosthetic setting,  Xu \etal\ describe a walking aid robot that was designed to be able to autonomously adapt to different users (Xu \etal\ 2013). The robot uses reinforcement learning to adjust the relative control of the human director in real time for smoother, faster movement. In this scenario, to ensure the safety of the user, the agency of the robot changes automatically to assume velocity control in some situations. Smoothness of motion, security of the system, and and intuitive control can all be viewed as different capacity functions that are impacted by the adaptive nature of the assistant.

\subsection{Goals: Reward-Based Control}

Goal-seeking behaviour on the part of both the director and the assistant enables a more detailed progression of interactions than is possible with an adaptive, but not goal-seeking, assistant. What follows is one example of a full life-cycle of training of an assistive machine, where both the director (the human) and assistant (an advanced prosthesis) are goal-seeking agents, and where the director thinks of the prosthesis as a goal-seeking agent. We can imagine the following progression:

\begin{enumerate}

\item At the onset, the director can only provide reward signals indicating their approval or disapproval; no other signals have any agreed upon meaning.
\item Using rewards, the machine can learn a function that maps a person's face, body language, tone of voice, and other cues to a valuation that is grounded in cumulative reward (a {\em value function}, as detailed by Sutton and Barto (1998), and used in face valuing by Veeriah \etal\ (2016)); this value function is a form of communicative capital, and enables further training in a way that is less tedious than the consistent delivery of primary rewards.
\item Using values, the director teaches the assistant a language that may be used to instruct it at a low level---e.g., simple commands, body language, cues like pointing, and the basics of shifting between different modes or functions of a system.
\item Using these developed instructions, higher-level commands, cues, and shorthands can be established between the director and the assistant; these may be subsequently used to create more abstract and more comprehensive communicative capital and, as a result, realize more of the partnership's capacity.
\end{enumerate}

With this progression in mind, there are a variety of compatible ways to incorporate human knowledge into a control learning system, either prior to or during learning, that fall at different points along a progression from primary reward to high-level direction and demonstration (as summarized by Thomaz and Breazeal (2008), Chernova and Thomaz (2014), Pilarski and Sutton (2012), Amershi \etal\ (2014), and others).
Starting with the idea of training based on primary reward, as in the progression described above, Knox and Stone (2009) formally introduced the {\em Interactive Shaping Problem}, wherein an agent is acting in an environment and a human is observing the agent's performance and providing feedback to the agent such that the agent must learn the best possible way to act based on that feedback. The interactive shaping problem is closely related to the discussion of communicative capital, as it is a readily observable case of information sharing between two goal-seeking systems with a limited channel of communication---in this case, a reward signal delivered by the director to the assistant.

Goal-seeking behaviour in an assistant, and a corresponding change in the director to model their assistant as goal-seeking agent, strictly increases the maximum capacity of a partnership. A director's interactions with an assistant are further supported by a new channel of communication with defined semantics (i.e., reward) that allows the director to shape the assistant's behaviour in ways that are not possible for a adaptive but non-goal-seeking assistant. This new channel is integral to realizing capacity to deal with non-stationary tasks, changing problem domains, and novel environments. Providing the means by which to shape behavior can also reduce the amount of pretraining for the system, as interactions are now accompanied by subjective human feedback. Reward allows the human to shape the assistant to perform the task in a personalized, and situation-specific way---an adaptive goal-seeking agent has the ability to incorporate engineered knowledge, but continually move beyond it.

While reward-based training (interactive shaping) of machine learners is an active field of research, goal-seeking behaviour in prosthetic devices is a less explored area. Previous work by Pilarski \etal\  demonstrated how both predefined and human-delivered reward could be delivered to a goal-seeking assistant to gradually improve the control capabilities of a myoelectric control interface (Pilarski \etal\ 2013b, Pilarski \etal\ 2011). By using a goal-seeking reinforcement learning agent to control the joints of a prosthesis, informed by implicitly or explicitly acquired predictions about future movement, the director-assistant team was found to be able to progressively refine the simultaneous multi-joint myoelectric control of a robotic arm. In these studies by Pilarski \etal, the director's approval and disapproval was delivered by the director to the assistant with full knowledge of the assistant's learning capacity. Extensions of these initial studies to more complex settings have been made, and help begin to inform whether or not this form of goal-seeking by the assistant holds merit for daily-life myoelectric control by people with amputations (Mathewson and Pilarski 2016). Specifically, Mathewson and Pilarski show that there are advantages and disadvantages of incorporating simultaneous human control and feedback signals in the reward-based training of a myoelectric robot arm. Their results demonstrate that there is a potential increase in the capacity of a partnership by incorporating human-feedback, but that this increase may be paired with a corresponding increase in demands on the director's cognition.

\subsection{Future work: Communication, models, and minds}

 An agent is only able to intentionally communicate about something that it is aware of. An agent may be aware of information coming to it from the environment, e.g., external information relating to sensation and sensorimotor signals. An agent may also be aware of signals originating from within itself (Sherstan \etal\  2016). These internal and external signals can form the basis for more abstract concepts or beliefs about how the world operates. Beliefs about the nature of internal and external signals are a kind of knowledge that we here broadly denote as {\em models}. For example, a prosthetic assistant may be able to represent a concept such as ``tea cup''. That concept may be thought of as a compressed model of the internal and external signals available to the assistant. It is then possible for the assistant to communicate that conceptual knowledge to another agent (e.g., the director). Higher-level representations of the world may be seen as a form of compression, and serve as the basis for communication. In order to achieve more effective communication (and form more comprehensive types of communicative capital) it is useful for an assistant to have or generate models of its partner and the world. Further, for a prosthetic assistant to develop a representation of the director's goals, it may be necessary to first be able to represent such goals internally.  As well described by Pezzulo \etal\ (2011), shared representations may be a critical part of communication during director-assistant interaction, and central to the formation of more effective models in terms of beliefs, actions, and intentions. This naturally takes us towards developing a \textit{theory of mind} (TOM)---an agent predicting the internal beliefs, motivations, and thoughts of another (especially as applied to observable sensorimotor interactions; Pezzulo \etal\ (2011 \& 2013), and Candidi \etal\ (2015)). In order to do this, a prosthetic assistant may first need to be capable of holding such models about itself and the director.

Models as they apply to a prosthetic director-assistant partnership may take many forms, from the mathematical to the empirical, where a first example of the latter would be large collections of learned, temporally extended predictions (e.g., Sutton \etal\ 2011, Pilarski and Sherstan 2016). Models that impact the capacity of a director-assistant partnership further include those that approximate the way the world operates (e.g., the transition between states or the rewards expected from a state transition, c.f., Sutton and Barto 1998). A goal-seeking assistant may also require models that begin to approximate the behaviour of the director, especially the behaviour in terms of how and when reward and other communications with agreed upon meanings are delivered. In this line of thinking, Knox \etal\ (2009, 2013, 2015) have provided one approach for a machine learner to build up an understanding of how humans deliver reward, and Vien and Ertel  (2013) have shown that human-feedback models can be generalized to address the problems associated with periods of noisy, or inconsistent, human feedback. This latter work may allow for the acquisition of more reliable communicative capital. Recent advancements in modeling human feedback with a Bayesian approach (ISABL or Inferring Strategy Aware Bayesian Learning) have further improved on the work of Knox and Stone in discrete environments (Loftin \etal\ 2014).

As one example of how models can impact a director-assistant partnership, Bicho \etal\ (2011) describe a human-robot shared construction task in which a humanoid robot and a human must work together to assemble a toy. Since their workspaces do not overlap, the task requires actions (i.e., reaching and grasping) from both the director and assistant. The robot assistant infers the goal of the human from contextual clues and acts accordingly, communicating its intention at each point during the task using a speech synthesizer. This allows the director to further model the internal processes of the assistant. Another example of a joint task in which a robot infers the goal of the director comes from Liu \etal\ (2016). In their work, participants and virtual robots collaborate to accomplish a task, and the robot infers the human's goal based on motion. This research suggests that goal inference (i.e., the modeling goals) decreased the time required to finish tasks and improved other measures of performance, including director-assistant trust.

In terms of the bidirectional communication that is supported by more advanced models, as we increase the agency of the assistant side of a partnership we increase the diversity and amount of information the assistant is able to supply (and choose to supply) to the director. In our primary example of an amputee and their prosthetic limb, we would usually refer to the information flowing from the prosthetic assistant to the human director as \textit{feedback}. With increasing agency on the prosthetic side, this feedback has the potential to contain more than just direct sensory information. An adaptive or goal-seeking assistant is also able to communicate what it has learned about the environment, the task, or the director---e.g., the prediction-based feedback demonstrated by Parker \etal\ (2014) and Edwards \etal\ (2016).

The impact of feedback from an adaptive prosthetic assistant is quantified in work by Parker \etal\ (2014). In their work, three different kinds of feedback were used to supply a director with information about how best to control the movements of a wearable robot in the form of a supernumerary limb (Fig.\ \ref{fig:examples}c)---no feedback, mechanistic feedback, and adaptive feedback in the form of predictions (with the last two cases representing the agency combinations in Fig. \ref{fig:combs}a and Fig. \ref{fig:combs}b respectively). The director needed to move the robotic assistant in a confined work space, coming as close as possible to the work space's walls without making physical contact. The director was blindfolded and was acoustically isolated by way of noise-canceling headphones, so that they only received information about the world via the assistant's feedback (in effect reducing the agency contributions from the director). 

\begin{figure}[t]
\centering
\includegraphics[height=1.9in]{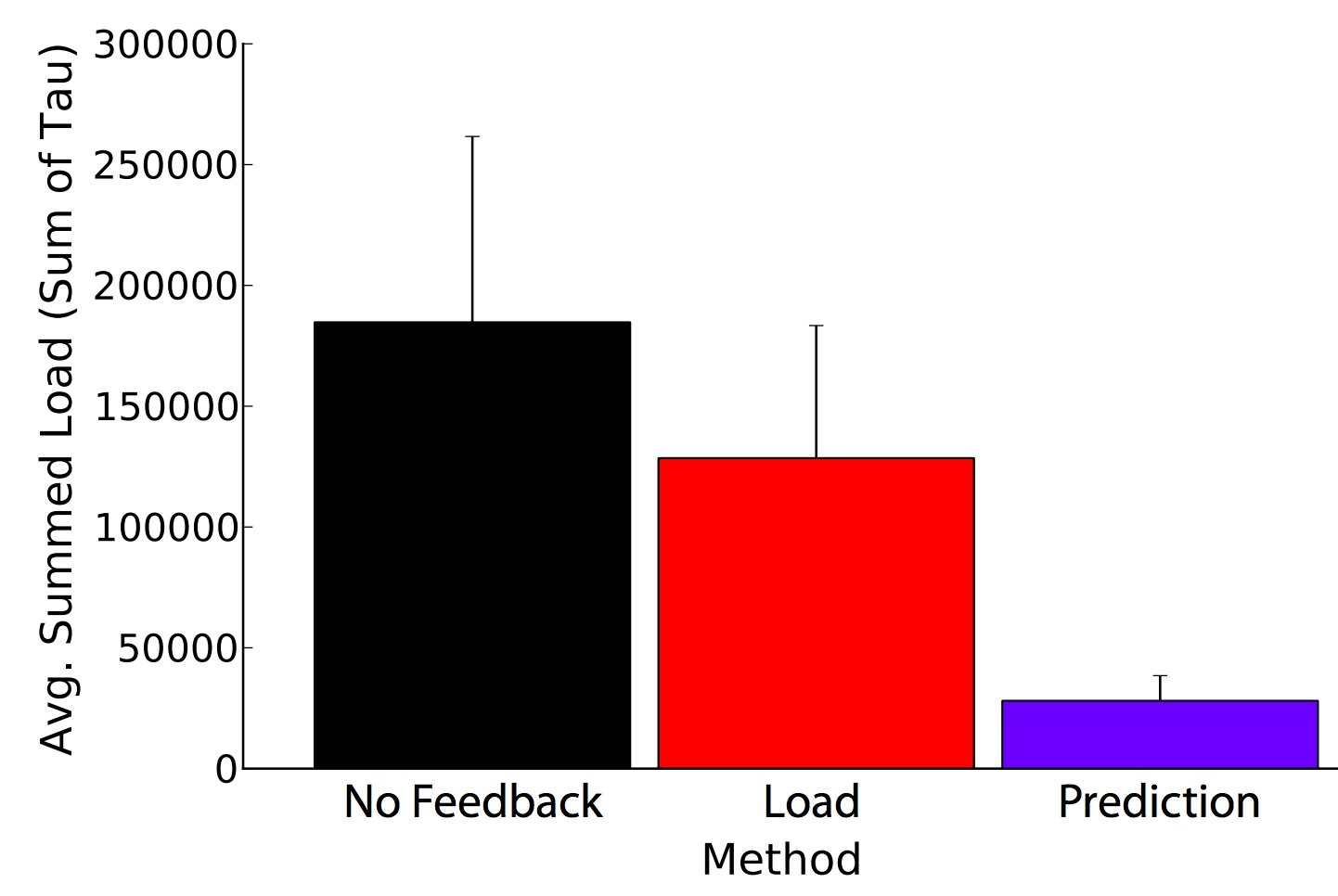}\hfill
\includegraphics[height=1.9in]{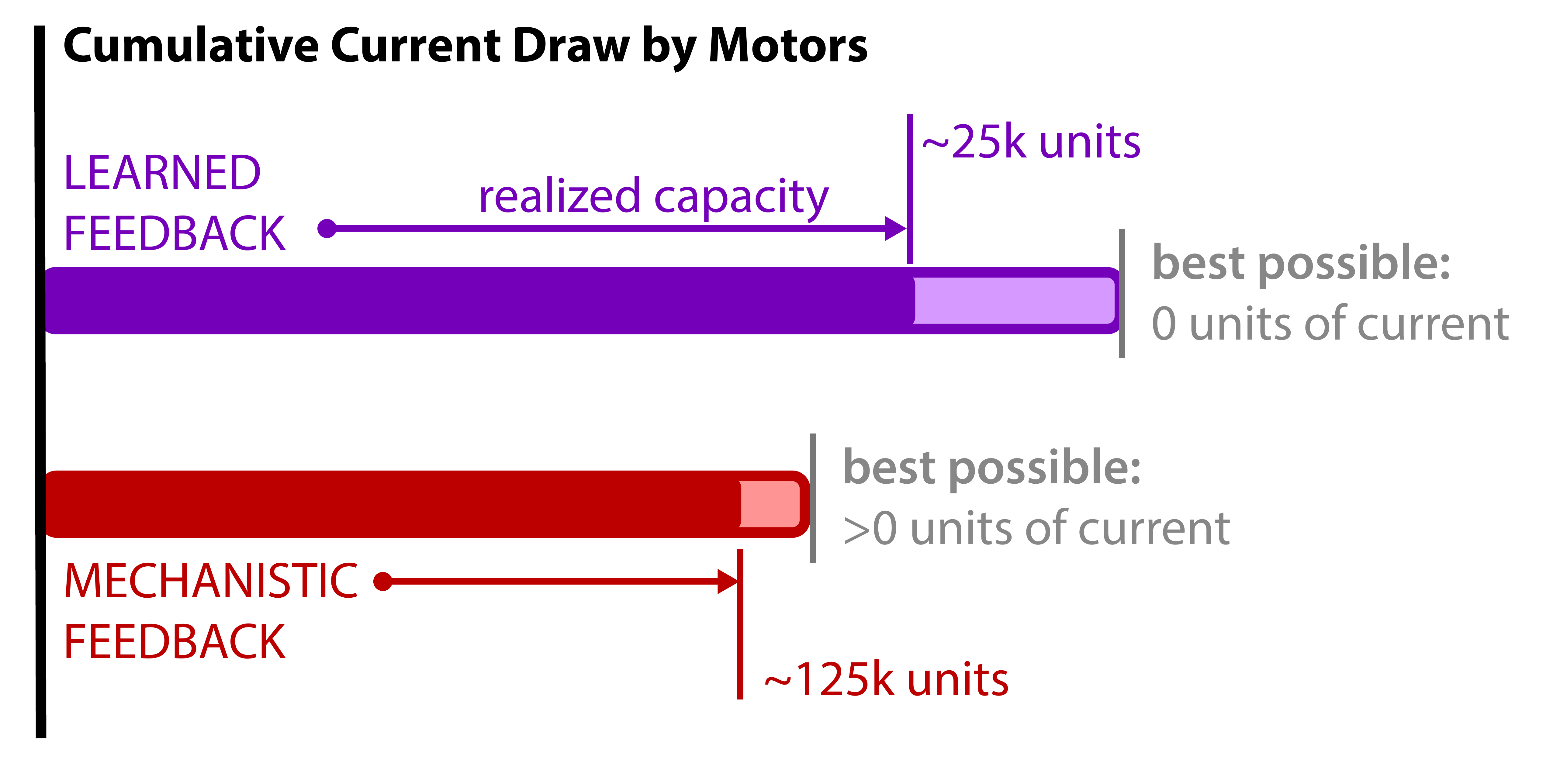}\\
\bf (a)\hfil (b)
\caption{A comparison of the difference between adaptive and mechanistic delivery of feedback by an assistant to a director during their control of a supernumerary limb (i.e., Fig. \ref{fig:examples}c). (a) The adaptive assistant realized significantly more capacity in terms of reducing the current drawn by the motors of the robot arm (plot adapted from Parker \etal\ 2014). (b) In this feedback setting, the agency of the director was controlled, i.e., the director did not know whether their assistant was adaptive or mechanistic. The increase in agency on the part of the assistant was shown to improve realized capacity of the partnership, and additional agency fundamentally improves the maximum possible capacity in terms of both current draw and impacts with the task work space.}
\label{fig:adamcap}
\end{figure}

The two capacity functions of interest in Parker \etal\ (2014) were as follows:  minimizing the current drawn by the motors due to impacts with the work space walls, and maximizing the number of times the director was able to use the arm to fully traverse the work space in the given time. On different trials, feedback from the device was either absent, delivered mechanistically upon contact with the walls, or delivered proportional to learned predictions about impacts with the walls. Realized capacity in terms of current draw was found to increase for the case where the director was paired with the adaptive assistant, but was found to approach a reduced maximum capacity for the case of mechanistic feedback from the assistant. A representation of these findings can be seen in Fig.\ \ref{fig:adamcap}---the current draw by the motors was dramatically reduced as the agency of the assistant was increased.
The results from Parker \etal\ (2014) also showed an increase in the number of times the arm traversed the work space as the agency of the assistant increased. Most of the discussion so far has focused on communicative capital that is related to the control of an assistant by a director. The work of Parker \etal\ (2014) therefore provides evidence as to how models and agency in the delivery of feedback (agency in bidirectional communication) may also help to realize the full capacity of a prosthetic partnership.

\section{Discussion and Related Work}

This article has so far discussed the setting of human-machine interaction, specifically the interactions between a human and their prosthetic technology. However, the ideas presented above regarding agency and communicative capital can be identified and analyzed in the interactions between any two or more intelligent systems. In this section we explore in detail both biological and non-biological examples of how agency plays a role in the interactions of multiple agents to achieve a goal; agency is examined through a discussion of seeing eye dogs, split-brain patients, intelligence amplification, and the human instruction of learning machines. We finish by connecting these examples back to the setting of human-prosthesis interaction, and, in the section that follows, propose an approach to testing the hypothesis that assistive devices can and should be thought of as systems with agency. The present work is also closely related to direct sensorimotor communication between human teams and partnerships. For more detail on the theory and practice of human-human interaction and sensorimotor communication during joint tasks, we refer the reader to representative articles by Wolpert \etal\ (2003), Sebanz \etal\ (2006, 2009), Pezzulo \etal\ (2011, 2013), and Candidi \etal\ (2015), and the related work of these authors.

\subsection{Seeing Eye Dogs}
A seeing-eye dog is perhaps the oldest example of an assistive technology with agency, with the earliest possible representation coming from the wall of a house excavated in Pompeii from c. 79 CE (Fishman 2003). Over time, seeing-eye dogs have come to be not only accepted, but viewed with esteem (Fishman 2003). Puppies are selected for the traits necessary in a successful guide dog. An example of the training of a guide dog, as done by ``The Seeing Eye", is discussed in Pfaffenberger \etal\ (1976). Volunteer homes do the initial obedience and house training before the dogs are given to the training school's instructors who have a knowledge of the psychology of dogs and the blind. It is crucial to ensure that the puppy is not overtrained before being transferred from the volunteer home to the instructor. The guide dog needs to be part of an active partnership---it must have the capacity to willingly disobey an instruction when it perceives a danger. {\em The party in charge of the interaction needs to be able to change from moment to moment in order for the partnership to be effective.}

A guide dog learns very specific behaviors and responses to commands which are not the normally desired responses from similar instructions given to other dogs. When the blind person drops something or issues a fetch command for example, the seeing eye dog is expected to retrieve the object, approach the blind person's right side, circle to the left side, and give an indication that is has the object. While being taken on normal training walks, a guide dog is expected to lead and tug slightly, which is contrary to the normal behavior we attempt to train into most dogs.

Because of these desired but atypical behaviors, the future owner must also be trained. This is another part of the instructor's job---they not only train the dog but the human director as well. The sighted instructor who has knowledge of both agents in the system is considered to be absolutely crucial to the success of the pairing. The blind person must be taught not only the precise vocabulary understood by the guide dog, but what to expect in response. Ultimately, the goal is for the dog to become an extension of the human partner. This requires both parties, human and dog, director and assistant, to learn each others' habits and idiosyncrasies in order to approach an ideal partnership.

In the seeing-eye-dog case, the instructor has communicative capital with both the dog and the human. The dog is well trained, but in a way differs from the normal human expectation, and certainly has agency. In fact, when the seeing eye dog lacks agency (is too obedient), it is considered a failure. The human user must learn how to effectively utilize the agency of the assistive technology (the seeing-eye dog). Time is taken from the human's normal life to train and to interact with the dog. The instructor uses the communicative capital it has with both director and assistant to facilitate, and possibly increase, the rate at which the director and assistant build capital with each other.

\subsection{Split-brain Patients}
The ideas of developing prostheses as goal-seeking intelligent agents and of communicative capital have close parallels to the research done on split-brain patients (Gazzaniga 2015, the comprehensive reference for this section unless otherwise noted). The corpus callosum serves as the primary communication channel between both hemispheres of the cortex. In split-brain patients the corpus callosum is surgically severed, eliminating the direct neural communication between the two cortices. This procedure has been used as an effective treatment for people with intractable epilepsy.

The splitting of the brain produces a situation which might be thought of as two goal-seeking agents occupying a single body, working together to accomplish tasks, and communicating across a channel with severely limited bandwidth---a situation very similar to one where a human interacts with an intelligent prosthesis. The lateral specialization of the brain becomes apparent when the two halves are severed. All information from the right visual field goes to the left hemisphere, while all information from the left visual field goes to the right hemisphere. Further, most somatosensory information and motor control is processed by the contra-lateral hemisphere. That is, the right hemisphere is primarily concerned with the left half of the body and vice versa.

Despite the severing of direct cortical intercommunication, split-brain patients often find unconscious ways of communicating between the two halves of their brains. They employ a phenomena known as cross-cueing, wherein one half of the brain uses the body and the environment as an indirect means of communication. For example, in one experiment a number was presented to the right hemisphere (the left visual field), and the patient was required to say the number. However, speaking requires use of the left hemisphere, making this task impossible without some form of communication between both halves of the brain; however, patients were able to do the task after some time. It was observed that when the number was presented, some physical action, such as head nods, would be used to count the number. The right hemisphere observed the number and nodded the head that many times. The left hemisphere counted the head nods and spoke the number aloud. In the terminology of this paper, this head nodding is an example of communicative capital---the two halves of the brain developed a communication protocol between themselves. What is truly fascinating is that the left-hemisphere, the speaking hemisphere, seems completely unaware that this is happening.

Further, communication of goals to both hemispheres is not straightforward. The majority of language processing and production is limited to the left hemisphere, although the right hemispheres of some patients appear to have limited abilities to understand language and/or read. Yet, patients learn to accomplish goals in a coordinated manner. This is done by using both cross-cueing and by interpreting context and the environment---an approach to prosthesis control that we have advocated for in the present work.

Finally, one of the open concerns about prostheses as intelligent goal-seeking agents is whether or not a person will accept an action they did not directly command. This is particularly of concern in the control of prosthetic arms. One hypothesis to emerge from split-brain research is the notion that the left hemisphere of our brains acts as an interpreter, rationalizing and providing stories about our current situation based on perceptual information, memory and knowledge of the world (Gazzaniga 2000). This interpreter is particularly evident in split-brain patients, where the left, speaking hemisphere will make up stories to explain the behavior of the left hand, which is guided by right hemisphere. Gazzaniga (2000) gives one example where the left hemisphere (right visual field) was presented with an image of a chicken claw and the right (left visual field) was presented with an image of snow. The subject then had to select associated images which were scattered on the table. The left hand selected a shovel and the right selected a chicken. When asked why the pair was chosen ``his left hemisphere replied `Oh that's simple. The chicken claw goes with the chicken, and you need a shovel to clean out the chicken shed'''.  It is possible that this same interpreter may enable an amputee to perceive the actions of a prosthetic device as their own. 

Further, it is fascinating that after splitting the callosum, patients are largely unaware of any changes. One would expect that upon losing all information about the left visual hemisphere, seeing only the right side of a person's face, for example, they would notice a gap in their perception. Instead, patients report that nothing has changed. Cross-cueing and the operation of the left side of the body, including perception and action, have fallen into the categories of unconscious processes. The left-brain interpreter no longer receives the signals which would convey that there was any visual information to be missed. This is similar to a condition known as anosognosia, where patients do not believe that the left hand side of their body is their own (Gazzaniga 2000). This condition is due to damage in the parietal cortex preventing communication of cortical sensory information from the right hemisphere to the left. In the case of amputation, we would expect these communication pathways to be intact and reporting a loss of sensory information to the interpreter. Thus, it seems likely that sensory feedback, of some kind, may be required to appease the interpreter. This may not require full restoration of natural sensation. It may be sufficient to use sensory substitution (Bach-y-Rita and Kercel 2003) or osseointegration (Oritz-Catalan 2014, Lundberg \etal\ 2011).

Taken as a whole, the work on split-brain patients can provide both motivation and understanding for how two intelligent goal-seeking agents can collaborate to control a single system.

\subsection{Intelligence amplification}

\textit{Intelligence Amplification} (IA) is the idea of using technology to increase a human's cognitive capacity. While formally investigated as early as the first half of the 20th century by Norbert Wiener, W. Ross Ashby, J. C. R. Licklider, Douglas Engelbart, and others, IA is not a new concept. Rather, IA is something humans have been doing for thousands of years. One can argue that the goal of most  existing technology, even language itself, is to enhance human cognitive capacities. This can take many forms including: improving our memory through the use of written language or computer databases, searching for information on our personal mobile devices, synthesizing and filtering information through the use of data tables or cutting-edge data analysis algorithms, and providing contextual information such as summarized records or augmented reality displays. As stated by Ashby (1956), amplifying the ability to select or choose between one of many options amplifies intellect, and this selection builds on a framework of two systems with a communication channel open between them. 

Clark (2015) has argued that our cognitive processes naturally extend into our environment and incorporate functional blocks of computation as available, even preferring those outside of our brains when they offer some benefit over internal ones. As an example, consider the following conversation between the historian Charles Weiner and physicist Richard Feynman. Upon finding some of Feynman's original notes, Weiner had commented that the materials represented ``a record of [Feynman's] day-to-day work." Feynman felt that it was, in fact, the cognitive process itself:

\begin{displayquote}
``I actually did the work on the paper,'' he said.\\
``Well,'' Weiner said, ``the work was done in your head, but the record of it is still here.''\\
``No, it's not a \textit{record}, not really. It's \textit{working}. You have to work on paper and this is the paper. Okay?'' (From Clark (2008), originally from Gleick (1993)).
\end{displayquote}

The question of this section then, is, how do the ideas of intelligent goal-seeking agents and communicative capital apply to IA? Consider the domain of computer interfaces. Computers, whether desktops, tablets, or smartphones, all augment our cognitive abilities. At present, there is significant effort to develop virtual assistants on such devices. Such assistants may have some level of agency; these assistants may be adaptive, changing their behavior and suggestions to meet the user's needs (as reviewed by Markoff (2015)). To date, existing computer interfaces have largely remained fixed and unadaptive. However, thanks in part to increases in available computation, computers are now improving in their ability to anticipate user needs and to provided users with the information and interfaces that are most needed at any given moment (Langley 1997, Markoff 2015). As shown by recent industrial interest in machine learning methods like reinforcement learning, approaching such an interface is theoretically possible by creating a goal-seeking agent that learns directly from how you interact with it.

Technologies that provide IA become even more interesting, and potentially powerful, when tightly coupled to the user. Many augmenting technologies, such as computers, are loosely coupled. That is, they are not always available as a cognitive resource. Arguably, smartphones are becoming increasingly coupled to their human users. Even more tightly coupled are body-worn prostheses and neuroprostheses. Such devices are nearly always available and physically coupled to the user. This tight coupling provides the brain with the greatest opportunity to adapt, via neuroplastic processes, to the device and truly integrate it into its cognitive flow. Such integration, often termed {\em embodiment}, has the potential to significantly reduce cognitive load and processing time.

Perhaps the split-brain research can again shed light on the idea of IA. In split-brain patients, it is typically observed that the left, communicative hemisphere is dominant and more cognitively capable (Gazzaniga 2015). The right hemisphere has goals that may not necessarily be directly in-line with those of the left hemisphere. Here we may consider the left-hemisphere to play the role of the user (director) and the right hemisphere to play the role of the  neuroprosthesis (assistant). 

The right hemisphere has limited capacity to comprehend and communicate verbal information. However, some patients do develop limited abilities to speak from the right hemisphere in single word utterances. In one example, a picture is presented to the right hemisphere and the left hemisphere is asked to describe the image. The researchers observed that the subject was able to describe the image through an iterative process. That is, they would make broad general statements and then refine them. How could this be? The left hemisphere should not have any information about the image. What the researchers deduced was that the left hemisphere was verbally making broad statements, which the right hemisphere could hear and understand. At key points the right hemisphere would interject with a word to fill in the missing information and the left hemisphere would continue to refine the description it was making based on that information. This is a clear example of one goal-seeking agent assisting another to perform a cognitive task. What is very interesting is that these patients seem to be largely unaware that they are \textit{cheating}, rather, the interpreter in the left hemisphere sees the behavior as part of themselves, part of their whole. Again, with a tightly coupled neuroprosthesis we might expect the brain to integrate it fully into its representation of self.

While we have focused on viewing the user (left hemisphere) and the neuroprosthesis (right hemisphere) as a system, there are, in fact, cognitive benefits to decoupling that occurs when the callosum is severed. In some types of tasks each hemisphere is able to process information on its own and operate independently, enabling the individual to perform tasks which are not possible for persons with intact callosum. For example, in one experiment subjects are shown one shape in each visual field and asked to draw the shape with the corresponding hand. That is, the left hand is to draw the shape on the left and right is to draw the shape on the right. When the shapes are the same both persons with intact callosum and those whose callosa are sectioned are able to perform the task. When the shapes are different, say a triangle in one and a circle in the other, only those with sectioned callosum are able to draw the shapes at the same time (Gazzaniga 2015). These cases demonstrate what is possible when an intelligent system is amplified by communication and coupling to another intelligent system (i.e., IA).

\subsection{Interactive approaches to instruction, communication and control}

There are multiple ways that a human and an agent---e.g., an assistive robot like a prosthesis---can beneficially interact to achieve the human's objectives (Argall \etal\ 2009, Pilarski and Sutton 2012, Thomaz and Breazeal \etal\ 2008). A pertinent family of methods, broadly classified as interactive machine learning, has demonstrated the potential to increase the capabilities of decision making systems in complex, dynamic, and novel environments (i.e., those common to prosthetic director-assistant interactions). In much of the existing interactive machine learning literature, one or more feedback channels are used as a means by which a non-expert can train, teach, and interact with a system without explicitly programming it. This shaping allows for the human to learn how the system accepts and interprets feedback, and for the system to learn what the human's goals are for their shared interaction (Thomaz \etal\ 2008).

Interactive machine learning has produced a number of important milestones. With repect to goal-driven systems, trial-and-error machine learning has been shown to be accelerated through the presentation of human-delivered reward and forms of intermediate reinforcement. Examples include the use of shaping signals (Kaplan \etal\ 2002), the delivery of reward from both a human and the environment (Knox and Stone 2012), multi-signal reinforcement (Thomaz and Breazeal 2008), and combinations of both direct control and reward-based feedback (Pilarski \etal\ 2011 \& 2013; Mathewson and Pilarski 2016).
As one example noted in Sec.\ 6 above, an agent's learning can be facilitated by a human host through interactive reinforcement learning (Knox \etal\ 2009, Knox \etal\ 2013, Knox and Stone 2012).
Griffith \etal\ (2013) built on the earlier work of Knox and Stone with a framework to maximize the information gained from human feedback. 
Loftin \etal\ (2014) have further expanded the space of human interaction through detailed investigation of human teaching strategies and developed systems which model the human feedback. Their systems have been shown to learn faster and with less feedback than other approaches. Interactive learning demonstrations and instructions have also been shown to help teach different ways of behaving to a learning machine (e.g., Argall \etal\ 2009, Chao \etal\ 2010; Judah \etal\ 2010; Kaplan \etal\ 2002; Lin 1991-1993). 
These human directions and examples supplement the signals already occurring in the sensorimotor stream of a machine learner.

As shown by the interactive machine learning studies described above, humans directors can utilize a number of different approaches to effectively communicate their subjective intentions and goals to machine learning assistants. 
Through interactive learning, information from a human director can help a machine learner to achieve arbitrary user-centric goals, can improve a system's learning speed, and can increase the overall performance of a learning system.
Advances in interactive machine learning therefore provide a basis for increasing the rate with which a prosthetic director-assistant partnership may realize capacity, and in certain cases can also be expected to extend the maximum capacity of a partnership.

\section{Steps to hypothesis testing}

Based on evidence to date, we expect that increasing the agency of a prosthetic assistant and progression of interactions will allow a collaborative partnership to ultimately accomplish tasks faster, easier, more safely, and more efficiently.  Work is now needed to aggressively test this hypothesis and identify, in concrete terms, the contributions and practical utility of agency and goal-seeking behaviour on the part of a prosthetic assistant.

\begin{figure}[t]
\centering
\includegraphics[height=3.0in]{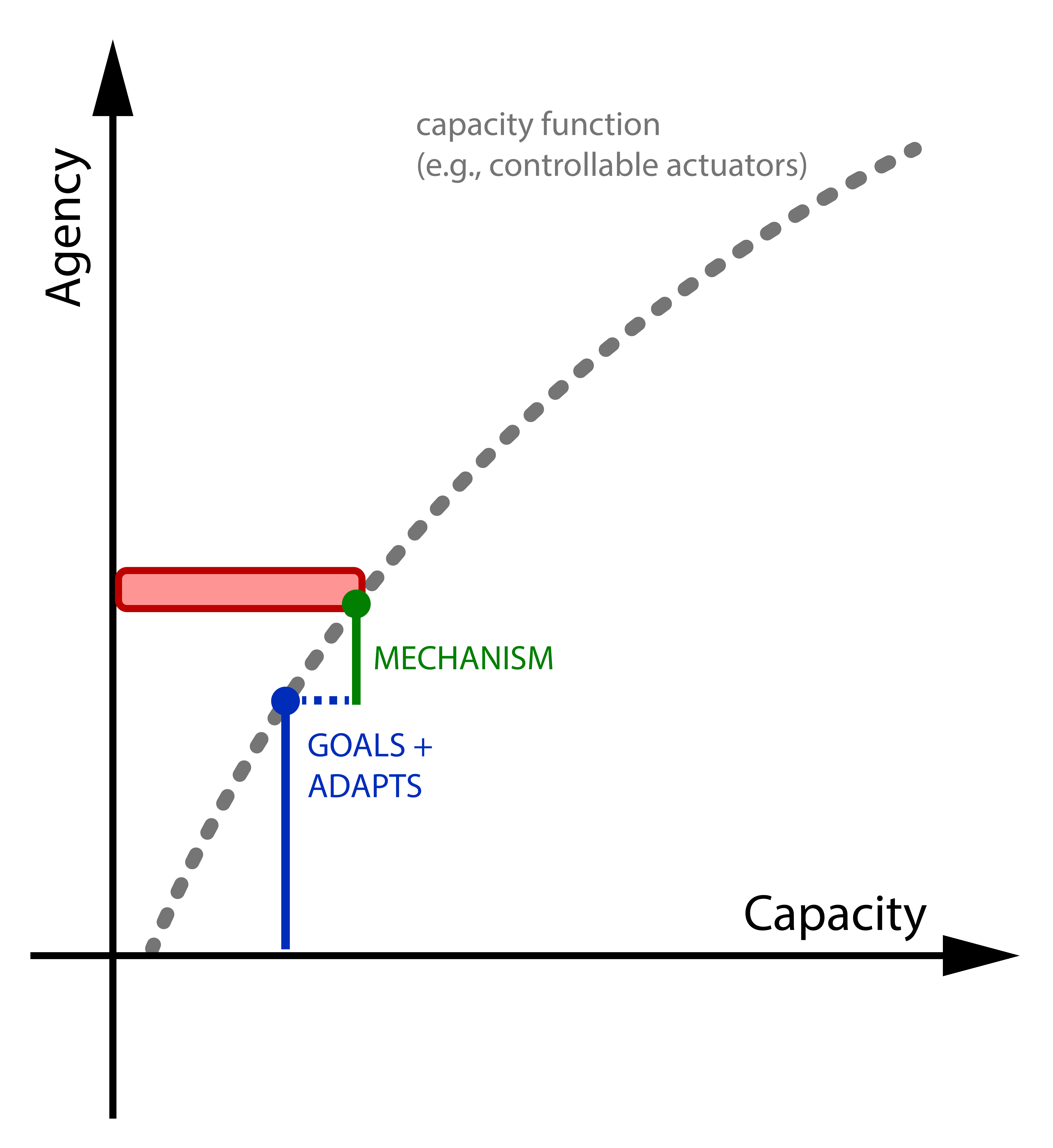}
\includegraphics[height=3.0in]{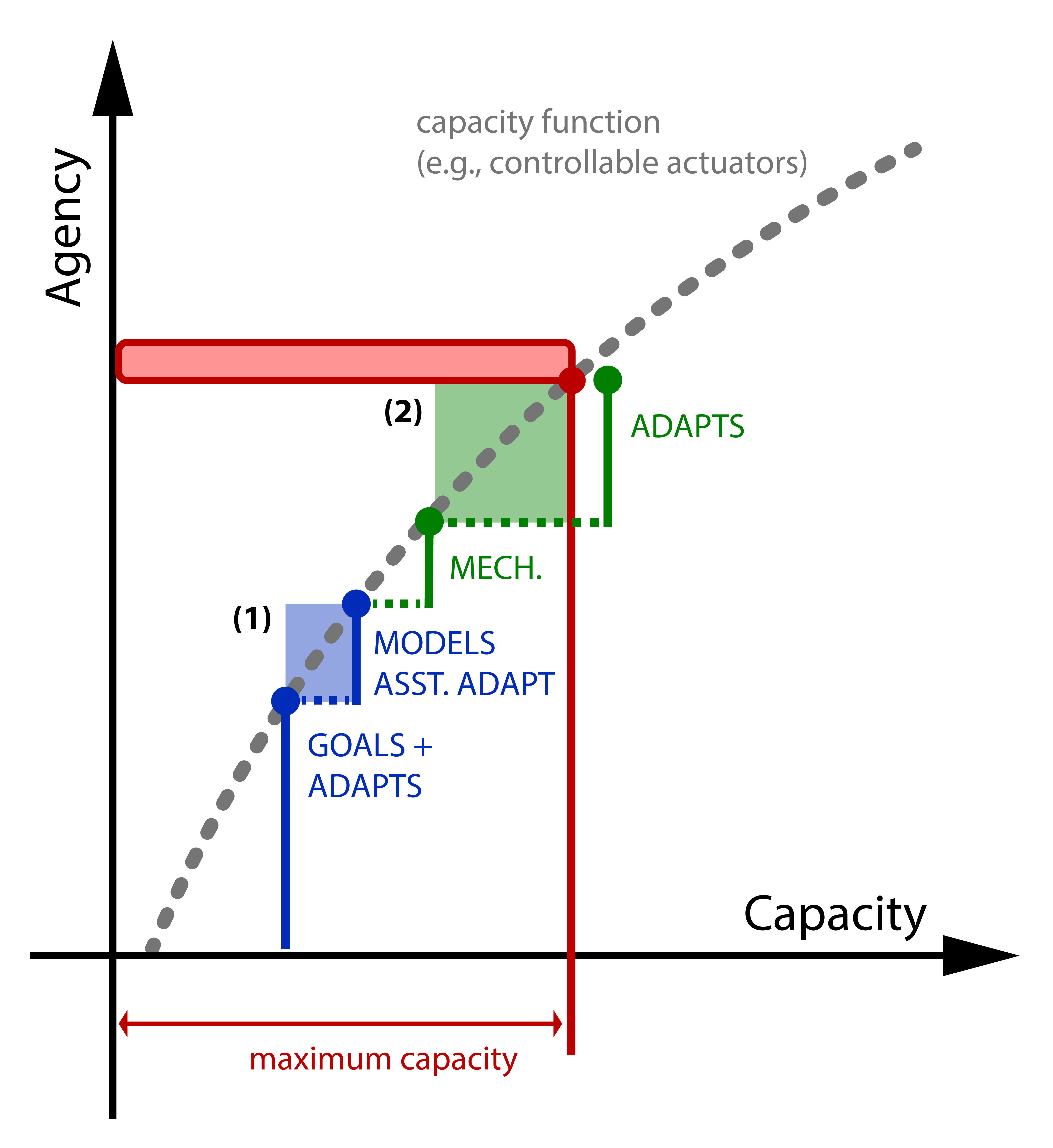}\\
\bf (a) \hfil (b)
\caption{Contributions from individual components of agency as they relate to the maximum capacity for a goal-seeking director paired with (a) a mechanistic assistant and (b) an adaptive assistant (director components shown in blue, and assistant components shown in green). Coloured rectangles (1) and (2) indicate the relative changes in agency and capacity for both the director and the assistant when the assistant is made able to adapt.}
\label{fig:testing}
\end{figure}

Figure \ref{fig:testing} uses our agency-capacity schema to compare a human-mechanism partnership to a partnership where the assistant is able to adapt (Fig.\ \ref{fig:testing}a, analogous to conventional myoelectric control, and Fig.\ \ref{fig:testing}b, analogous to pattern recognition or adaptive switching). As described in Sec. 6.2, increased agency on the part of the assistant enables increased agency on the part of the director. This increase is depicted in Fig.\ \ref{fig:testing}, where the relative changes in agency and capacity by both parties are shown via coloured rectangles (1) and (2). While it is clear from a broad body of experimental results and user reports that adaptive systems promise to better meet the needs of a prosthetic director, it is at present not clear how to precisely quantify the exact amount of capacity and agency that changes as we modify the operation of the director or the assistant---i.e., is is not clear how best to establish the width and height of rectangles (1) and (2) in Fig.\ \ref{fig:testing}b. 

To rigorously quantify the contributions of agency to human enhancement, and thus investigate the central hypothesis that prosthetic assistants should be full goal-seeking agents, we believe that the following questions must be addressed in the context of human-prosthesis interaction:

\begin{itemize}
\item {\bf How should we measure agency?} Is it possible and beneficial to create a measurable, smoothly varying scale of agency, or is it more valuable to think in terms of discrete blocks of agency that can be added or subtracted but are without concrete measure?
\item {\bf Can we precisely control the amount of agency in a both the director and assistant?} To provide strong evaluations of the impact of agency on capacity, it will be important to titrate the degrees of agency within a partnership and evaluate the resulting changes. Is rigorous titration possible? 
\item {\bf What is the best way to think of the qualitative contributions of different kinds of agency?} For example, what aspects of a partnership change when agency changes for one of the parties? As in Fig. \ref{fig:testing}, when the ability to adapt is added to an assistant, {\em both} the assistant and the director can be viewed as having increased agency.
\item {\bf How should we define the type and shape of capacity functions?} By defining capacity functions, we can quantify how capacity increases with agency---i.e., the horizontal part of rectangles (1) and (2) in Fig.\ \ref{fig:testing}. Is it possible to measure the shape of task-implicit capacity functions, and for those we define to facilitate our testing, what are the right capacity functions to use for individual tasks? Should we instead look to capacity functions that are independent of a given task or setting?
\item {\bf What is the right way to think about maximum and realized capacity?} In some cases it is straightforward to define maximum capacity and measure realized capacity, as it is implicit in the formulation of a task or setting. In other cases, maximum and realized capacity may only be inferred (or defined) by comparing a partnership to normative examples that we define as ideal.  What choices should be made if it is not possible to define maximum capacity?
\item  {\bf How should we quantify and study the rate with which capacity is realized?} Further, is it important to relate this rate in some way to the communicative capital that is and can be acquired by a partnership? For example, if only the director is realizing capacity, how does this compare to the rate of realized capacity when both the director and the assistant are building up communicative capital? We expect the latter case should realize capacity faster. However, experiments must be set up to also identify the potential for maladaptation, oscillation, or divergence.
\item {\bf What is the right way to visualize communicative capital?} Is it possible to precisely identify the communicative capital that is built up by the director and the assistant? In some cases we can identify learned models and other knowledge that is built during interaction, while in other cases it may be challenging or impossible to clearly isolate the capital that has been built. Is identifying specific communicative capital a requirement for studying the impact of agency on human enhancement, or is the idea of communicative capital useful primarily as a conceptual tool to help think about the way capacity is realized?
\end{itemize}

\subsection{Suggested experimental methodologies}

Taking into consideration the questions above, it is our recommendation that researchers begin to design experiments such that they are able to precisely control the agency of both the assistant {\em and the director}. Once designed, experiments should vary these levels of agency in predictable or controlled ways so as to quantify and qualitatively assess the contributions from each added component of agency. Assessment should be done according to a battery of capacity curves, some with known maximums (and measurable realized capacity) and some with empirically determined performance maximums.

A natural way to begin testing is through the conventional outcome measures used to assess the impact of rehabilitation interventions such as body-powered and myoelectric prostheses (c.f., Resnik \etal\ 2011 \& 2012; Hebert \etal\ 2009; Light \etal\ 1985). Outcome measures provide a clearly defined notion of capacity. Further, prosthetic outcome measures are already used to study the benefits of pairing patients to systems with different mechanistic levels of agency (e.g., during prosthetic fitting and patient assessment). In the majority of clinically deployed prostheses, the control approach and system design of the assistant is fixed. The communicative capital of the mechanism---how it interprets body signals and maps them to actuators---provides immediate realized capacity at a level determined by the mechanism's designers. Measures like the Southampton Hand Assessment Procedure, the Box-and-Blocks Task, and others are used to provide a quantitative assessment of the impact of these prosthetic mechanisms (Mathiowetz \etal\ 1985, Light \etal\ 2002). Rigorous, incremental testing of agency is therefore highly compatible with existing approaches, and will be significantly extended as more comprehensive motor, sensory, and cognitive outcome measures are developed.

One fruitful avenue for experimentation, as explored in Parker \etal\ (2014), is to deliberately reduce the agency of the director. Humans have a large degree of agency and this makes it difficult to assess the value of the much more limited agency of the prosthetic assistants we are currently able to develop and test. By reducing the agency of the human side of a partnership (e.g., by removing control options or sensory input to the director as they complete a task), we can see more clearly how different degrees of agency in the assistant contribute to the partnership, without the assistant's contribution being washed out by the director's contributions. A second, complementary view is to dramatically increase the agency of the assistant beyond what is in fact technically possible, so as to study the potential outcomes and sufficient conditions that support the collaboration of two systems with human-level agency. One way to do this is a type of sham trial known as a {\em Wizard-of-Oz} experiment (e.g., Viswanathan \etal\ 2014. We expect that results from prosthetic Wizard-of-Oz experiments should align with those from human-human sensorimotor collaboration (e.g., Candidi \etal\ 2015).

\newpage
\section{Conclusions}

In this work we presented the hypothesis that prosthetics and other technologies for human enhancement can and should be thought of as intelligent systems with agency. As a main contribution, we developed an agency-capacity schema for thinking about how components and degrees of agency impact the capacity of a human-prosthesis partnership.  Using this schema, we showed via examples from the published literature that increasing the agency of a prosthesis (the assistant) in many cases significantly improves the capabilities of its human user (the director). We draw three main conclusions from this work as contributions toward a new theory on human-prosthesis interaction: 1) that thinking of an assistive or augmentative device as a general goal-seeking agent clarifies the range of possibilities for robust and flexible interaction, 2) that, in particular, an agent-based viewpoint suggests a structured progression from primitive to more capable human-device interaction, and 3) that communicative capital, especially as implemented in the form of predictions, is a strong basis for progressive assistance and augmentation. The idea of communicative capital helps showcase commonalities in widely differing systems---as seen in examples from assistive animals, split-brain patients, and other forms of intelligence amplification, communicative capital is a useful way to study how different partnerships are effected and to compare these partnerships on even ground. Machine intelligence, and specifically machine learning, enables the acquisition and use of communicative capital to support a human-prosthesis partnership. In conclusion, we believe that an agency-based viewpoint on assistive technology contributes unique and complementary ideas to the future development of highly functional prosthetic devices. Further studies are now needed to examine the precise impact that differing forms of  agency have on prosthetic control.

\section*{Conflict of Interest Statement}

The authors declare that the research was conducted in the absence of any commercial or financial relationships that could be construed as a potential conflict of interest.

\section*{Author Contributions}

All authors contributed to the conception and design of the work, the interpretation of data and studies, drafting and revising the work for important intellectual content, the approval of the version to be submitted/published, and agree to be accountable for all aspects of the work.

\section*{Funding}
This research was undertaken, in part, thanks to funding from the Canada Research Chairs program, the Canada Foundation for Innovation, the Alberta Innovates Centre for Machine Learning / the Alberta Machine Intelligence Institute, Alberta Innovates -- Technology Futures, Google DeepMind, and the Natural Sciences and Engineering Research Council. 

\section*{Acknowledgments}
The authors thank the other members of the Bionic Limbs for Natural Improved Control Laboratory and the Reinforcement Learning and Artificial Intelligence Laboratory for many helpful thoughts and comments. They specifically thank Michael Rory Dawson, Jacqueline Hebert, and Jaden Travnik for a number of fruitful discussions and feedback.

\section*{References}

\parindent=0pt
\small
\def\hangin{\hangindent=0.15in}
\parskip=12pt

Amershi, S., Cakmak, M., Knox, W. B.,  Kulesza, T. (2014). Power to the people: The role of humans in interactive machine learning. {\em AI Magazine} 35 (4), 105--120. doi:10.1609/aimag.v35i4.2513	
 
Argall, B. D., Chernova, S., Veloso, M., Browning, B. (2009). A survey of robot learning from demonstration. \textit{Robotics and Autonomous Systems, 57} (5), 469--483. doi: 10.1016/j.robot.2008.10.024

Ashby, W. R. (1956). {\em An introduction to cybernetics}. London: Chapman \& Hall.

Bach-y-Rita, P., Kercel, S. W. (2003). Sensory substitution and the human-machine interface. \textit{Trends in Cognitive Sciences, 7} (12), 541--546. doi: 10.1016/j.tics.2003.10.013

Belfiore, M. P. (2009). {\em The department of mad scientists: how DARPA is remaking our world, from the internet to artificial limbs.} New York: Smithsonian Books; Harper.

Bicho, E., Erlhagen, W., Louro, L., Costa e Silva, E. (2011). Neuro-cognitive mechanisms of decision making in joint action: A human-robot interaction study. {\em Human Movement Science} 30, 846--868.

Brooks, R. A. (2002). {\em Flesh and machines: How robots will change us.} New York: Pantheon Books.

Candidi, M., Curioni, A., Donnarumma, F., Sacheli, L. M., Pezzulo, G. (2015). Interactional leader-follower sensorimotor communication strategies during repetitive joint actions. {\em J. R. Soc. Interface} 12, 20150644. doi:10.1098/rsif.2015.0644

Carmena, J.M. (2012). Becoming bionic. {\em IEEE Spectrum} 49 (3), 24--29. 

Castellini, C., Artemiadis, P., Wininger, M., et al. (2014). Proceedings of the first workshop on peripheral machine interfaces: Going beyond traditional surface electromyography. \textit{Frontiers in Neurorobotics, 8} (22), 1--17. doi:10.3389/fnbot.2014.00022

Chao, C., Cakmak, M., Thomaz, A.~L.\ (2010). Transparent active learning for robots. \textit{Proceedings of the 5th ACM/IEEE International Conference on Human-robot Interaction (HRI)}, March 2--5, Osaka, Japan, 317--324.

Chernova S., Thomaz, A. L. (2014). {\em Robot learning from human teachers.} Synthesis Lectures on Artificial Intelligence and Machine Learning, Morgan \& Claypool Publishers. doi:10.2200/S00568ED1V01Y201402AIM028

Clark, A. (2008). \textit{Supersizing the mind: Embodiment, action, and cognitive extension}. Oxford: Oxford University Press.

Clark, A. (2015). Embodied prediction. In: {\em Open Mind: Frankfurt am Main: MIND Group.} Metzinger, T. \& Windt, J. (eds.), 21 pages. doi: 10.15502/9783958570115

Dawson, M. R., Sherstan, C., Carey, J. P., Hebert, J. S., Pilarski, P. M. (2014). Development of the BENTO Arm: An improved robotic arm for myoelectric training and research. \textit{Proceedings of the Myoelectric Controls Symposium (MEC)}, August 18--22, Fredericton, Canada, 60--64.

Dewdney, C. (1998). {\em Last flesh: Life in the transhuman era}. Harper Collins Canada. 

Doidge, N. (2007). {\em The brain that changes itself: Stories of personal triumph from the frontiers of brain science}. New York: Viking.

Edwards, A. L., Dawson, M. R., Hebert, J. S., et al. (2015). Application of real-time machine learning to myoelectric prosthesis control: A case series in adaptive switching. {\em Prosthetics \& Orthotics International}. Published online before print September 30, 2015, doi:10.1177/0309364615605373.

Edwards, A. L., Hebert, J. S., Pilarski, P. M. (2016). Machine learning and unlearning to autonomously switch between the functions of a myoelectric arm. {\em Proceedings of the 6th IEEE RAS/EMBS International Conference on Biomedical Robotics and Biomechatronics} (BioRob2016), June 26-29, 2016, Singapore, 514--521.

Edwards, A. L. (2016). {\em Adaptive and autonomous switching: Shared control of powered prosthetic arms using reinforcement learning.} M.Sc. Thesis, University of Alberta.

Feil-Seifer, D., Mataric, M. J. (2005). Socially assistive robotics. \textit{Proceedings of the 9th International Conference on Rehabilitation Robotics (ICORR)}, June 28--July 1, Chicago, USA, 465--468. doi: 10.1109/ICORR.2005.1501143

Fishman, G. A. (2003). When your eyes have a wet nose: the evolution of the use of guide dogs and establishing the seeing eye. \textit{Survey of Ophthalmology} 48 (4), 452--458.

Gazzaniga, M. S. (2000). Cerebral specialization and interhemispheric communication: Does the corpus callosum enable the human condition? \textit{Brain : A Journal of Neurology 123} (7), 1293--326. doi: 10.1093/brain/123.7.1293

Gazzaniga, M. S. (2015). \textit{Tales from both sides of the brain: A life in neuroscience}. New York, NY: Ecco.

Geary, J. (2002). {\em The body electric: An anatomy of the new bionic senses}. New Brunswick, N.J: Rutgers University Press.

Gleick, J. (1993). \textit{Genius: The life and times of Richard Feynman}. Vintage.

Griffith, S., Subramanian, K., Scholz, J. (2013). Policy shaping: Integrating human feedback with reinforcement learning. \textit{Proceedings of the Advances in Neural Information Processing Systems (NIPS)}, December 5--10, Lake Tahoe, USA, 2625--2633.

Hebert, J. S., Wolfe, D. L., Miller, W. C., Deathe, A. B., Devlin, M., Pallaveshi, L. 2009. Outcome measures in amputation rehabilitation: ICF Body Functions. {\em Disability and Rehabilitation} 31 (19), 1541--1554.

Hebert, J. S., Elzinga, K., Chan, K. M., Olson, J., Morhart, M. (2014). Updates in targeted sensory reinnervation for upper limb amputation. \textit{Current Surgery Reports 2} (3), 1--9. doi: 10.1007/s40137-013-0045-7

Herr, H. (2009). Exoskeletons and orthoses: Classification, design challenges and future directions. \textit{Journal of Neuroengineering and Rehabilitation 6}(21). doi: 10.1186/1743-0003-6-21

Hochberg, L. R., Serruya, M. D., Friehs, G. M., et al. (2006). Neuronal ensemble control of prosthetic devices by a human with tetraplegia. \textit{Nature} 442 (7099), 164--171. doi: 10.1038/nature04970

Judah, K., Roy, S., Fern, A., Dietterich, T. G. (2010). Reinforcement learning via practice and critique advice. \textit{Proceedings of the 24th AAAI Conference on Artificial Intelligence}, July 11--15, Atlanta, USA, 481--486.

Kaplan, F., Oudeyer, P.~Y., Kubinyi, E., Miklosi, A. (2002). Robotic clicker training. \textit{Robotics and Autonomous Systems} 38, 197--206.

Knox, W. B., Stone P. (2009). Interactively shaping
agents via human reinforcement: The TAMER framework. In {\em 
Proc. of 5th Intl. Conf. on Knowledge Capture}, 9--16. ACM.

Knox, W.~B., Stone, P. (2012). Reinforcement learning from simultaneous human and MDP reward. In \textit{Proceedings of the 11th International Conference on Autonomous Agents and Multiagent Systems (AAMAS)}, June 4--8, Valencia, Spain, 475--482.

Knox, W. B., Stone, P., Breazeal, C. (2013). Training a robot via human feedback: A case study. {\em Proc. of the International Conference on Social Robotics}, 460--470. Springer International Publishing.

Knox, W. B., Stone P. (2015). Framing reinforcement learning from human reward: Reward positivity, temporal discounting, episodicity, and performance. {\em Artificial Intelligence} 225, 24--50.

Kollar, T., Tellex, S., Roy, D., Roy, N. (2010). Toward understanding natural language directions. \textit{Proceedings of the 5th ACM/IEEE International Conference on Human-robot Interaction (HRI)}, March 2--5, Osaka, Japan, 259--266. doi: 10.1109/HRI.2010.5453186

Langley, P. (1997). Machine learning for adaptive user interfaces. In {\em  KI-97: Advances in artificial intelligence}, 53--62. Springer Berlin Heidelberg.

Light, C. M., Chappell, P. H., Kyberd, P. J. (2002). Establishing a standardized clinical assessment tool of pathologic and prosthetic hand function: normative data, reliability, and validity. {\em Arch.\ Physical Medicine and Rehabilitation} 83, 776-783.

Lin, L.-J. (1991). Programming robots using reinforcement learning and teaching. In \textit{Proceedings of the 9th National Conference on Artificial Intelligence (AAAI)}, July 14--19, Anaheim, USA, 781--786.

Lin, L.-J. (1992). Self-improving reactive agents based on reinforcement learning, planning and teaching. \textit{Machine Learning} 8, 293--321.

Lin, L.-J. (1993). Hierarchical learning of robot skills by reinforcement. In \textit{Proceedings of the International Joint Conference on Neural Networks (IJCNN)}, October 25--29, Nagoya, Japan, 181--186.

Liu, C., Hedrick, J. K. (2016). Cooperative search using human-UAV teams. {\em AIAA Infotech @ Aerospace, AIAA SciTech}, AIAA 2016-1653.
doi:10.2514/6.2016-1653 

Loftin, R. T., MacGlashan, J., Peng, B., Taylor, M. E., Littman, M. L., Huang, J., and Roberts, D. L. (2014). A strategy-aware technique for learning behaviors from discrete human feedback. In {\em AAAI}, 937--943.

Lundberg, M., Hagberg, K., Bullington, J. (2011). My prosthesis as a part of me: a qualitative analysis of living with an osseointegrated prosthetic limb. \textit{Prosthtetics and Orthotics International} 35 (2), 207--214. doi: 10.1177/0309364611409795.

Markoff, J. (2015). \textit{Machines of loving grace: The quest for common ground between humans and robots}. Ecco

Markovic, M., Dosen, S., Cipriani, C., Popovic, D., Farina, D. (2014). Stereovision and augmented reality for closed-loop control of grasping in hand prostheses. \textit{Journal of Neural Engineering} 11 (4), 046001. doi: 10.1088/1741-2560/11/4/046001

Mathiowetz, V., Volland, G., Kashman, N., Weber, K. (1985). Adult norms for the box and block test of manual dexterity. {\em The American Journal of Occupational Therapy} 39 (6), 386--391.

Mathewson, K. W., Pilarski, P. M. (2016). Simultaneous control and human feedback in the training of a robotic agent with actor-critic reinforcement learning. In: {\em 25th International Joint Conference on Artificial Intelligence, Interactive Machine Learning Workshop}, July 2016, New York, USA.

Micera, S., Carpaneto, J., Raspopovic, S. (2010). Control of hand prostheses using peripheral information. \textit{IEEE Reviews in Biomedical Engineering} 3, 48--68.

Mill\'an, J. d. R., Rupp, R., Müller-Putz, G. R., Murray-Smith, R., et al. (2010). Combining brain-computer interfaces and assistive technologies: State-of-the-art and challenges. \textit{Frontiers in Neuroscience} 4 (161). doi: 10.3389/fnins.2010.00161

Moon, A., Parker, C. A. C., Croft, E. A., Van der Loos , M. H. F. (2013). Design and impact of hesitation gestures during human-robot resource conflicts. {\em Journal of Human-Robot Interaction} 2 (3), 18--40. doi: 10.5898/JHRI.2.3.Moon

Moss, F. (2011). {\em The sorcerers and their apprentices: How the digital magicians of the MIT media lab are creating the innovative technologies that will transform our lives.} New York: Crown Business.

Office of the Secretary, United States Department of Health, Education and Welfare (1979). Belmont Report: Ethical principles and guidelines for the protection of human subjects of research, report of the national commission for the protection of human subjects of biomedical and behavioral research. {\em Federal Register} 44 (76), 23191--7.

Ortiz-Catalan, M., Hakansson, B., Branemark, R. (2014). An osseointegrated human- machine gateway for long-term sensory feedback and motor control of artificial limbs. {\em Sci Transl Med} 6 (257), 257re6. doi:10.1126/scitranslmed.3008933

Parker, A. S. R., Edwards, A. L., Pilarski, P. M. (2014). Using learned predictions as feedback to improve control and communication with an artificial limb: Preliminary findings. arXiv:1408.1913 [cs.AI], 1--7.

Parker, P., Englehart, K., Hudgins, B. (2006). Myoelectric signal processing for control of powered limb prostheses. \textit{Journal of Electromyography and Kinesiology} 16 (6), 541--548. doi:10.1016/j.jelekin.2006.08.006

Peerdeman, B., Boere, D., Witteveen, H., Huis in `t Veld, R., Hermens, H., et al.  (2011). Myoelectric forearm prostheses: State of the art from a user-centered perspective. \textit{The Journal of Rehabilitation Research and Development} 48 (6), 719--738. doi:10.1682/JRRD.2010.08.0161

Pezzulo, G.,  Dindo, H. (2011). What should I do next? Using shared representations to solve interaction problems. {\em Exp Brain Res}  211, 613--630. doi:10.1007/s00221-011-2712-1

Pezzulo G., Donnarumma F., Dindo H. (2013). Human sensorimotor communication: A theory of signaling in online social interactions. {\em PLoS ONE} 8 (11), e79876. doi:10.1371/journal.pone.0079876

Pilarski, P. M., Dawson, M. R., Degris, T., Fahimi, F., Carey, J. P., Sutton, R. S. (2011). Online human training of a myoelectric prosthesis controller via actor-critic reinforcement learning. \textit{Proceedings of the 2011 IEEE International Conference on Rehabilitation Robotics (ICORR)}, June 29--July 1, Zurich, Switzerland, 134--140. doi:10.1109/ICORR.2011.5975338

Pilarski, P., Sutton, R. (2012). Between instruction and reward: Human-prompted switching. \textit{2012 AAAI Fall Symposium Series}, 46--52.

Pilarski, P. M., Dawson, M. R., Degris, T., Carey, J., Chan, K. M., Hebert, J. S., Sutton, R. S. (2013a). Adaptive artificial limbs: A real-time approach to prediction and anticipation. \textit{IEEE Robotics and Automation Magazine} 20 (1), 53--64. doi: 10.1109/MRA.2012.2229948

Pilarski, P. M., Dick, T. B., Sutton, R. S. (2013b). Real-time prediction learning for the simultaneous actuation of multiple prosthetic joints. \textit{Proceedings of the 13th IEEE International Conference on Rehabilitation Robotics (ICORR)}, June 24--26, Seattle, USA,  doi: 10.1109/ICORR.2013.6650435

Pilarski, P. M., Sutton, R. S., Mathewson, K. W. (2015). Prosthetic devices as goal-seeking agents. {\em Second Workshop on Present and Future of Non-Invasive Peripheral-Nervous-System Machine Interfaces: Progress in Restoring the Human Functions} (PNS-MI), Singapore, Aug. 11, 2015, 4 pages.

Pilarski, P. M., Sherstan, C. (2016). Steps toward knowledgeable neuroprostheses. {\em Proceedings of the 6th IEEE RAS/EMBS International Conference on Biomedical Robotics and Biomechatronics} (BioRob2016), June 26-29, 2016, Singapore, pp. 220. 

Pfaffenberger, C. J., Scott, J. P., Fuller, J. L., Ginsburg, B. E., Biefelt, S. W. (1976). \textit{Guide dogs for the blind: Their selection, development, and training}. Amsterdam, The Netherlands: Elsevier Scientific Publishing Company.

Rashidi, P., Mihailidis, A. (2013). A survey on ambient-assisted living tools for older adults. \textit{IEEE Journal of Biomedical and Health Informatics} 17 (3), 579--590. doi: 10.1109/JBHI.2012.2234129

Resnik, L., Meucci, M. R., Lieberman-Klinger, S., Fantini, C., Kelty, D. L., Disla, R., Sasson, N. (2012). Advanced upper limb prosthetic devices: Implications for upper limb prosthetic rehabilitation. \textit{Archives of Physical Medicine and Rehabilitation} 93 (4), 710--717. doi: 10.1016/j.apmr.2011.11.010

Resnik, L. 2011. Development and testing of new upper-limb prosthetic devices: Research designs for usability testing. {\em JRRD} 48 (6), 697--706.

Saridis, G. N., Stephanou, H. E. (1977). A hierarchical approach to the control of a prosthetic arm. \textit{IEEE Transactions on Systems, Man and Cybernetics} 7 (6), 407--420. doi: 10.1109/TSMC.1977.4309737

Scheme, E., Englehart, K. B. (2011). Electromyogram pattern recognition for control of powered upper-limb prostheses: state of the art and challenges for clinical use.\textit{Journal of Rehabilitation Research and Development} 48 (6): 643--660.

Schofield, J. S., Evans, K. R., Carey, J. P., Hebert, J. S. (2014). Applications of sensory feedback in motorized upper extremity prosthesis: A review. \textit{Expert Review of Medical Devices} 11 (5), 499--511. doi: 10.1586/17434440.2014.929496

Sebanz N, Bekkering H, Knoblich G (2006) Joint action: Bodies and minds moving together. {\em Trends. Cogn. Sci.} 10 (2), 70--76.

Sebanz N, Knoblich G (2009) Prediction in joint action: What, when, and where. {\em Top. Cogn. Sci.} 1, 353--367.

Sherstan, C., Modayil, J., Pilarski, P. M. (2015). A collaborative approach to the simultaneous multi-joint control of a prosthetic arm. \textit{In International Conference on Rehabilitation Robotics (ICORR)}, August 11--14, Singapore, Singapore, 13--18.

Sherstan, C., Machado, M. C., White, A., Pilarski, P. M. (2016). Introspective agents: Confidence measures for general value functions. {\em Proc. of the Ninth Conference on Artificial General Intelligence} (AGI-16), New York City, July 16-19, {\em Lecture Notes in Computer Science} (9782), 258--261. Springer.

Sutton, R. S., Barto, A. G. (1998). \textit{Reinforcement learning: An introduction}. Cambridge, MA: MIT Press. doi: 10.1109/TNN.1998.712192

Sutton, R. S., Modayil, J., Delp, M., Degris, T., Pilarski, P. M., White, A., Precup, D. (2011). Horde: A scalable real-time architecture for learning knowledge from unsupervised sensorimotor interaction. \textit{Proceedings of the 10th International Conference on Autonomous Agents and Multiagent Systems (AAMAS)}, May 2--6, Taipei, Taiwan, 761--768). doi: 10.1037/a0023964

Thomaz, A. L., Breazeal, C. (2008). Teachable robots: Understanding human teaching behaviour to build more effective robot learners. \textit{Artificial Intelligence} 172 (6--7), 716--737. doi: 10.1016/j.artint.2007.09.009

Tosic, P. T., Agha, G. A. (2004). Towards a hierarchical taxonomy of autonomous agents. {\em Proc. 2004 IEEE International Conference on Systems, Man and Cybernetics} (4), 3421--3426. IEEE.

Veeriah, V., Pilarski, P. M., Sutton, R. S. (2016). Face valuing: Training user interfaces with facial expressions and reinforcement learning. \textit{2016 IJCAI Workshop on Interactive Machine Learning}. arXiv:1606.02807

Vien, N. A., Ertel, W., Chung, T. C. (2013). Learning via human feedback in continuous state and action spaces. {\em Applied intelligence} 39 (2), 267--278.

Viswanathan, P., Bell, J. L., Wang, R. H. L., Adhikari, B., Mackworth, A. K., Mihailidis, A., … Mitchell, I. M. (2014). A wizard-of-oz intelligent wheelchair study with cognitively- impaired older adults: Attitudes toward user control. \textit{Proceedings of IEEE/RSJ International Conference on Intelligent Robots and Systems (IROS),} September 14--18, Chicago, USA.

Williams, T. W. (2011). Progress on stabilizing and controlling powered upper-limb prostheses. \textit{Journal of Rehabilitation Research and Development} 48 (6), ix--xix. doi: 10.1682/JRRD.2011.04.0078

Wolpert, D. M., Doya, K., Kawato, M. (2003) A unifying computational framework for motor control and social interaction. {\em Philos.\ Trans.\ R.\ Soc.\ Lond.\ B.\ Biol.\ Sci.} 358 (1431), 593--602.

Xu, W., Huang, J., Wang, Y., Cai, H. (2013). Study of reinforcement learning based shared control of walking-aid robot. {\em Proceedings of the 2013 IEEE/SICE International Symposium on System Integration}, Kobe, Japan.

Ziegler-Graham, K., MacKenzie, E. J., Ephraim, P. L., Travison, T. G., Brookmeyer, R. (2008). Estimating the prevalence of limb loss in the united states: 2005 to 2050. \textit{Archives of Physical Medicine and Rehabilitation} 89 (3), 422--429. doi: 10.1016/j.apmr.2007.11.005

Zuo, K. J., Olson, J. L. (2014). The evolution of functional hand replacement: From iron prostheses to hand transplantation. \textit{Canadian Journal of Plastic Surgery} 22 (1), 44--51.

\end{document}